\documentclass[twoside]{article} 
\usepackage[accepted]{aistats2024}

\usepackage{algorithm}
\usepackage{algorithmicx}
\usepackage{algpseudocode}
\usepackage{amsmath, amsthm}
\usepackage{comment}
\usepackage{amssymb}
\usepackage{dsfont}
\usepackage{graphicx}
\usepackage{relsize}
\usepackage{nicefrac}
\usepackage{subcaption}
\usepackage{xcolor}
\usepackage{pgffor} 
\definecolor{citecolor}{RGB}{0,128,128}
\usepackage{placeins}
\usepackage{hyperref}
\hypersetup{colorlinks,linkcolor=citecolor,citecolor=citecolor,urlcolor=citecolor}  

\usepackage{cleveref}

\DeclareMathOperator*{\argmin}{arg\,min}

\newcommand{\R}{\mathbb{R}}

\newcommand{\allocore}{\textsc{Al$\ell_0$core}}
\newcommand{\titleallocore}{AL$\ell_0$CORE}

\newcommand{\Ytensor}{\mathbf{Y}}
\newcommand{\yentry}{\textrm{y}_{\obs}}

\newcommand{\bi}{\textbf{d}}
\newcommand{\bk}{\textbf{k}}
\newcommand{\obs}{\bi} 
\newcommand{\lat}{\bk} 

\newcommand{\dm}[1]{\textrm{d}_{#1}}
\newcommand{\im}{\textrm{d}_m}
\newcommand{\kq}{\bk_q}
\newcommand{\kqm}{\textrm{k}_{q,m}}

\newcommand{\Pois}[1]{\textrm{Pois}\left( #1 \right)}

\newcommand{\phimdk}[3]{\phi_{#2, #3}^{\mathsmaller{(#1)}}}
\newcommand{\phimk}[2]{\boldsymbol{\phi}_{#2}^{\mathsmaller{(#1)}}}

\usepackage[round]{natbib}

\begin{document}

\twocolumn[
\aistatstitle{The AL$\ell_0$CORE Tensor Decomposition for Sparse Count Data}
\aistatsauthor{John Hood \And Aaron Schein}
\aistatsaddress{University of Chicago \And  University of Chicago} ]

\begin{abstract}
This paper introduces \allocore, a new form of probabilistic tensor decomposition. \allocore~is a Tucker decomposition that constrains the number of non-zero elements (i.e., the $\ell_0$-norm) of the core tensor to be at most $Q$. While the user dictates the total budget $Q$, the locations and values of the non-zero elements are latent variables allocated across the core tensor during inference. \allocore---i.e., \underline{allo}cated $\underline{\ell_0}$-\underline{co}nstrained~\underline{core}---thus enjoys both the computational tractability of canonical polyadic (CP) decomposition and the qualitatively appealing latent structure of Tucker. In a suite of real-data experiments, we demonstrate that \allocore~ typically requires only tiny fractions (e.g.,~1\%) of the core to achieve the same results as Tucker at a correspondingly small fraction of the cost.~\looseness=-1
\end{abstract}

\section{INTRODUCTION}
 
Tensor decomposition methods~\citep{kolda_tensor_2009,cichocki_tensor_2015} aim to find parsimonious representations of multi-mode data by seeking constrained (e.g.,~low-rank or sparse) reconstructions of observed tensors. When well-tailored, such methods are effective at distinguishing salient structure in data from idiosyncratic noise. Tensor decomposition methods have proven useful for a variety of predictive tasks like de-noising \citep{marot2008advances}, imputation~\citep{tomasi2005parafac}, and forecasting~\citep{xiong2010temporal}, as well as for counterfactual prediction in the context of causal modeling~\citep{amjad_mrsc_2019}. Due to their parsimony, tensor decomposition methods also tend to be highly interpretable, and are routinely used for exploratory and descriptive data analysis.

The two predominant forms of tensor decomposition, canonical polyadic (CP) and Tucker decomposition, each have benefits and drawbacks. CP decomposes an observed tensor $\Ytensor$ of shape $D_1 \times \dots \times D_M$ into $M$ factor matrices $\Phi^{\mathsmaller{(1)}}, \dots, \Phi^{\mathsmaller{(M)}}$, the $m^{\textrm{th}}$ having shape $D_m \times K$. CP can be understood as learning $K$ latent \textit{classes}, the $k^{\textrm{th}}$ of which represents a particular multi-mode pattern in the data via the factor vectors $\phimk{1}{k}, \dots \phimk{M}{k}$. This decomposition is simple and computationally advantageous, scaling generically with $K$ and the size of the observed tensor, $|\Ytensor| = \prod_{m=1}^{M} D_m$, as $\mathcal{O}(|\Ytensor| \,\cdot\, K)$.\looseness=-1

The Tucker decomposition, on the other hand, seeks a richer representation. It allows for different numbers of factors in each mode, such that the $m^{\textrm{th}}$ factor matrix $\Phi^{\mathsmaller{(m)}}$ takes shape $D_m \times K_m$. Each latent class then corresponds to one of the $\prod_{m=1}^M K_m$ unique combinations of per-mode factors, and is weighted by its corresponding entry in the $K_1 \times \dots \times K_M$ \textit{core tensor} $\boldsymbol{\Lambda}$. This richer representation, while qualitatively appealing, enacts a price: what is known as Tucker's ``exponential blowup'' is the fact that inference scales with the size of the core, $|\boldsymbol{\Lambda}|=\prod_{m=1}^M K_m$, as $\mathcal{O}(|\Ytensor| \cdot |\boldsymbol{\Lambda}|)$.

\begin{figure}[t]
    \centering
    \begin{subfigure}[b]{0.15\textwidth}
        \centering
        \includegraphics[width=\textwidth]{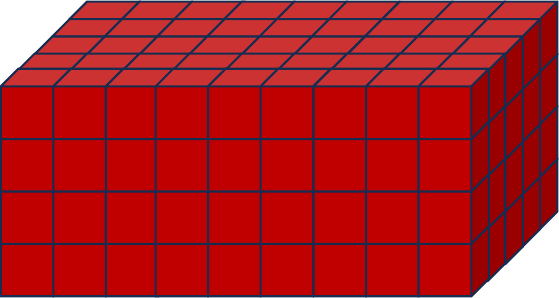}
        \caption{Tucker}
        \label{fig:tucker_core}
    \end{subfigure}
    \hfill % This adds a horizontal space between the two subfigures.
    \begin{subfigure}[b]{0.15\textwidth}
        \centering
        \includegraphics[width=\textwidth]{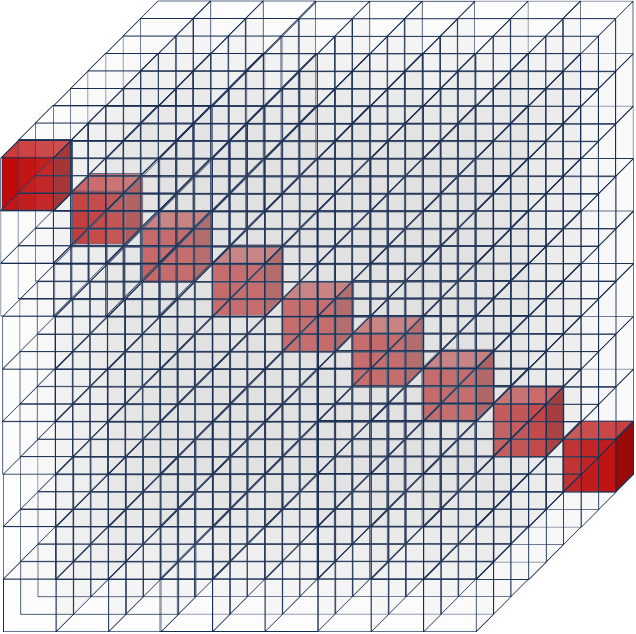}
        \caption{CP}
        \label{fig:cp_core}
    \end{subfigure}
    \hfill % This adds a horizontal space between the two subfigures.
    \begin{subfigure}[b]{0.15\textwidth}
        \centering
        \includegraphics[width=\textwidth]{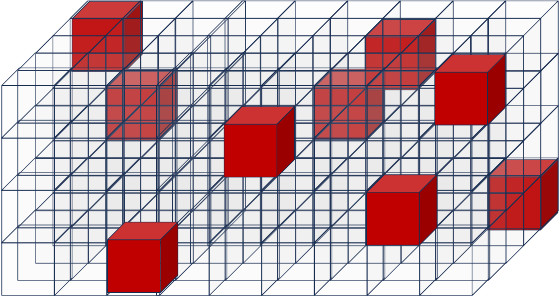}
        \caption{\allocore}
        \label{fig:allocore_core}
    \end{subfigure}
    \caption{The core tensor $\boldsymbol{\Lambda}$ in three related tensor decompositions. Transparent versus red values denote zero versus non-zeros. \allocore~relies on sparsity to achieve the representational richness of Tucker without suffering its ``exponential blowup'' in parameters.} 
    \label{fig:cores}
\end{figure}

This paper seeks a middle ground between CP and Tucker.  We introduce a new family of probabilistic tensor decomposition models that enjoy the representational richness of Tucker without its attendant computational cost. Our approach is inspired by ideas in \textit{count} tensor decomposition, and tailored accordingly.

Many tensors in practice are count tensors, or contingency tables, which arise naturally as summaries of tabular data. Such tensors tend to be very large but very sparse---i.e., $|\Ytensor| \gg ||\Ytensor||_0$---where the $\ell_0$-norm counts the number of non-zeros. Many tensor decomposition methods are specifically tailored to this setting and adopt loss functions or likelihoods consistent with the assumption that the observed entries are Poisson-distributed. Under a Poisson assumption, CP generally scales only as  $\mathcal{O}(||\Ytensor||_0 \cdot K)$, and Tucker as $\mathcal{O}(||\Ytensor||_0 \cdot |\boldsymbol{\Lambda}|)$, both of which represent a substantial computational improvement over their generic counterparts, particularly when $\Ytensor$ is very sparse.

Several recent papers connect Tucker decomposition under a Poisson assumption to models of complex networks~\citep{schein_bayesian_2016,de_bacco_community_2017,aguiar_tensor_2023}. Tucker has particular qualitative appeal in this setting, allowing for the representation of overlapping communities and cross-community interactions within different network layers and over different temporal regimes. Complex networks are often encoded as very large, sparse count tensors, and thus Poisson Tucker's dependence on only $||\Ytensor||_0$ is critical.~\looseness=-1

Despite this, Tucker's qualitative appeal still remains incompatible with its computational cost. Modelers often seek to represent large numbers of latent communities, particularly in cases where observed networks contain large numbers of actors. However, specifying large numbers of communities drives the ``exponential blowup'' in Tucker's core tensor and is thus often infeasible. As a result, modelers typically only use Tucker decomposition with small core tensors. Tucker's conceptual appeal for the analysis of complex networks, among other applications involving large sparse tensors, is therefore not easily attainable, in practice.\looseness=-1

This paper breaks ground on the analysis of large sparse tensors with the \allocore~(pronounced ``al-oh-core") decomposition. \allocore~is a Tucker decomposition where the core tensor is constrained to have at most $Q$ non-zero entries---i.e., $||\boldsymbol{\Lambda}||_0 \leq Q$. The budget $Q$ can be fixed in advance, tuned as a hyperparameter, or inferred as a latent variable. The locations and values of the non-zero entries are treated as latent variables and allocated across the core during inference---the name ``\underline{allo}cated \underline{$\ell_0$}-\underline{co}nstrained \underline{core}'' is intended to emphasize this central idea. With this constraint,~\allocore~can explore the same exponentially large latent space as Tucker, but scale generically with only $||\boldsymbol{\Lambda}||_0$; see~\cref{fig:cores}. When tailored to count tensors, where it is particularly motivated, \allocore~scales with $\mathcal{O}(||\Ytensor||_0 \cdot ||\boldsymbol{\Lambda}||_0)$, and thus exploits the sparsity in both high-dimensional observed and latent spaces.

\pagebreak

One might wonder why such a simple idea was not introduced sooner. Indeed, the literature on matrix and tensor (including Tucker) decompositions with both soft (e.g., $\ell_1$) and hard (e.g., $\ell_0$) sparsity constraints is vast. As we show, the computational advantage of the \allocore~decomposition's sparsity constraint is unlocked when combined with a sampling-based approach to inference, which iteratively re-samples the coordinates of the non-zero locations from their complete conditionals. This approach both ensures that instantiations of the core tensor are sparse and permits model-fitting in sub-exponential time. Previous work that has sought to improve the computational efficiency of Tucker has typically looked to optimization, EM, or variational inference (VI)-based approaches, which are commonly believed to be faster than MCMC. Such approaches however, if na\"ively adopted to infer the non-zeros in the core, would suffer the same ``exponential blowup'' as full Tucker. The computational advantages of the \allocore~decomposition are naturally tied to MCMC, which we speculate is the reason it has thus far been overlooked.~\looseness=-1

An outline of the paper is as follows. We introduce notation and provide background on tensor decomposition in~\cref{sec:bg}. We then sketch a general recipe for \allocore~in~\cref{sec:allocore}, which can be combined with a variety of different probabilistic assumptions. In~\cref{sec:bpallocore}, we give a complete example of such a model, by building Bayesian Poisson~\allocore, which is tailored to sparse count tensors representing dynamic multilayer networks. We further provide a full Gibbs sampling algorithm for this model. In~\cref{sec:connections}, we review and draw connections to related work, highlighting other tensor decomposition methods that are either tailored to sparse data, like complex networks, or incorporate sparsity constraints. In \cref{sec:realdata}, we then report a suite of experiments on real data of international relations, where we show that~\allocore~can achieve substantially better predictive performance than full Tucker at only a tiny fraction of the cost, while yielding rich and interpretable latent structure.

\section{PRELIMINARIES}
\label{sec:bg}

\textbf{Basic notation and terminology.} we use lowercase symbols for scalars $\textrm{y} \in \mathbb{R}$, lowercase bold symbols for vectors $\textbf{y} \in \mathbb{R}^{D}$, uppercase symbols for matrices $\textrm{Y} \in \mathbb{R}^{D_1 \times D_2}$, and uppercase bold symbols for tensors $\textbf{Y} \in \mathbb{R}^{D_1 \times \dots \times D_M}$.  A tensor $\Ytensor$ of shape $D_1 \times \dots \times D_M$ has $M$ \textit{modes}, with the $m^{\textrm{th}}$ mode having dimensionality $D_m$. The tensor contains scalar entries $\yentry$ where $\obs \equiv (\dm{1},\dots,\dm{M})$ is a \textit{multi-index} containing $M$ coordinates that collectively identify a cell or location in the tensor. The $m^{\textrm{th}}$ coordinate can take $D_m$ values, denoted as $d_m \in [D_m] \equiv \{1,\dots, D_m\}$.

\textbf{Tensor decomposition.} Tensor decompositions seek a \textit{reconstruction} $\widehat{\Ytensor}$---informally, $\Ytensor \approx \widehat{\Ytensor}$---that is a deterministic function of model parameters $\widehat{\Ytensor}\equiv \widehat{\Ytensor}(\Theta)$. Non-probabilistic methods seek to minimize a loss function $\widehat{\Theta} \leftarrow \argmin_{\Theta} \ell(\Ytensor, \widehat{\Ytensor}(\Theta))$, while probabilistic methods define a likelihood $P(\Ytensor \mid \widehat{\Ytensor})$ and perform MLE, or introduce priors and perform posterior inference. Such likelihoods typically factorize---i.e.,
\begin{equation}
P(\Ytensor \mid \widehat{\Ytensor}) = \prod_{\obs} P(\yentry \mid \widehat{\textrm{y}}_{\obs}).
\end{equation}
\textbf{Poisson tensor decomposition.} For sparse count tensors, it is common to assume a Poisson likelihood---i.e., $P(\yentry \mid \widehat{\textrm{y}}_{\obs}) = \Pois{\yentry;\, \widehat{\textrm{y}}_{\obs}}$. Among other demonstrated benefits, this assumption leads to inference algorithms that scale only with the number of non-zeros $||\Ytensor||_0$, which can be seen by writing the terms of the log-likelihood proportional to model parameters,
\begin{align*}
\log P(\Ytensor \mid \widehat{\Ytensor}) &= \log \prod_{\obs} \Pois{\yentry;\, \widehat{\textrm{y}}_{\obs}} \\
&\propto_{\widehat{\Ytensor}} \sum_{\obs} \yentry\log \widehat{\textrm{y}}_{\obs} - \sum_{\obs} \widehat{\textrm{y}}_{\obs},
\end{align*}
and observing that summands in the first sum need only be calculated at the non-zeros $\yentry > 0$, while the second sum does not depend on the data and can typically be computed efficiently for standard parameterizations of $\widehat{\textrm{y}}_{\obs}$. Poisson tensor decomposition can be understood alternatively as minimizing the generalized KL loss or $\mathcal{I}$-divergence~\citep{chi_tensors_2012}. Its computational advantage applies both to explicitly probabilistic approaches, even those reliant on full Bayesian inference, as well as to non-probabilistic ones. We note that all parameters must be \textit{non-negative}, to satisfy the definition of the Poisson, which promotes a ``parts-based'' representation~\citep{lee_learning_1999} that is highly interpretable.

\textbf{Tucker decomposition~\citep{tucker_mathematical_1966}.} Under the Tucker decomposition, the reconstruction $\widehat{\mathbf{Y}}$ is the following multilinear function of model parameters:
\begin{align}
\label{eq:tucker}
\widehat{\mathbf{Y}} = \sum_{\textrm{k}_1=1}^{K_1} \dots \sum_{\textrm{k}_M=1}^{K_M} \lambda_{\textrm{k}_1,\dots,\textrm{k}_M} \,  (\boldsymbol{\phi}^{\mathsmaller{(1)}}_{\textrm{k}_1} \circ \dots \circ \boldsymbol{\phi}^{\mathsmaller{(M)}}_{\textrm{k}_M} )
\end{align}
Unpacking this equation, each term $\phimk{m}{\textrm{k}_m} \in \mathbb{R}^{D_m}$ represents the $(\textrm{k}_m)^{\textrm{th}}$ \textit{latent factor} in the $m^{\textrm{th}}$ mode, which is stored as a column vector in the factor matrix $\Phi^{\mathsmaller{(m)}} \in \mathbb{R}^{D_m \times K_m}$. The notation $(\circ)$ denotes the outer product, and thus each term $(\boldsymbol{\phi}^{\mathsmaller{(1)}}_{\textrm{k}_1} \circ \dots \circ \boldsymbol{\phi}^{\mathsmaller{(M)}}_{\textrm{k}_M} )$ is a tensor of the same shape as $\Ytensor$---i.e., $D_1 \times \dots \times D_M$. Tucker can thus be viewed as a weighted sum of tensors, each weighted by $\lambda_{\textrm{k}_1,\dots,\textrm{k}_M}$, an entry in the $K_1 \times \cdots \times K_M$ \textit{core tensor} $\boldsymbol{\Lambda}$.  We will often adopt multi-index notation to denote locations in the core, such that an entry is $\lambda_{\lat}$, where $\lat \equiv (\textrm{k}_1,\dots, \textrm{k}_M)$.

\textbf{CP decomposition~\citep{harshman_foundations_1970}.} More widely used than Tucker is the canonical polyadic (CP) decomposition (also known as CANDECOMP or PARAFAC). It is a special case of Tucker, where the latent dimensions are all equal to a single value $K=K_1=\dots=K_M$ and the core tensor has only $K$ non-zero values $\boldsymbol{\lambda}=(\lambda_1,\dots,\lambda_K)$ which are arranged along its super-diagonal $\boldsymbol{\Lambda}=\textrm{diag}(\boldsymbol{\lambda})$---i.e.,
\begin{align*}
\lambda_{\textrm{k}_1,\dots,\textrm{k}_M} &= \begin{cases}
\lambda_k &\textrm{if } k=\textrm{k}_1=\dots=\textrm{k}_M\\
0 &\textrm{otherwise}
\end{cases}
\end{align*}
In such a setting, \cref{eq:tucker} can be simplified to the form usually given to describe CP---i.e.,
\begin{align}
\label{eq:cp}
\widehat{\mathbf{Y}} = \sum_{k=1}^{K} \lambda_k\,  (\boldsymbol{\phi}^{\mathsmaller{(1)}}_{k} \circ \dots \circ \boldsymbol{\phi}^{\mathsmaller{(M)}}_{k} )
\end{align}

\textbf{Tucker versus CP.} Both decompositions in~\cref{eq:tucker,eq:cp} can be viewed as weighted sums over a set of \textit{latent classes}. While CP infers $K$ classes, each defined by a unique set of $M$ factors $\phimk{1}{k},\dots,\phimk{M}{k}$, Tucker infers $\prod_{m=1}^M K_m$ classes, each defined by a \textit{unique combination} of factors $\phimk{1}{\textrm{k}_1},\dots,\phimk{M}{\textrm{k}_M}$. Tucker thus involves a much greater degree of parameter sharing, which can make it more statistically efficient, and does not require all modes have the same latent dimensionality, which can be qualitatively appealing. However, this all comes at a computational cost. Both decompositions scale with the number of latent classes, which for Tucker is exponential in the number of modes.
\section{\titleallocore~DECOMPOSITION}
\label{sec:allocore}
This section presents a general recipe for \allocore, which combines a sparsity constraint with a sampling-based approach to inference to simultaneously achieve the qualitative appeal and statistical efficiency of Tucker without suffering its computational cost.~\looseness=-1 

\allocore~is a Tucker decomposition (\cref{eq:tucker}) whose core tensor is constrained to have at most $Q$ non-zeros,
\begin{equation}\label{eq:l0bound}
||\boldsymbol{\Lambda}||_0 \leq Q
\end{equation}
where $Q \in \mathbb{N}$ is a positive integer-valued \textit{budget} which can be fixed in advance, tuned as a hyperparameter, or inferred as a latent variable. 

Unlike in CP decomposition, which also has a sparse core tensor, the non-zero entries in~\allocore~need not necessarily lie along the core tensor's super-diagonal, and the core tensor need not necessarily be a hypercube with $K_1=\dots=K_M$.~\looseness=-1

Instead, with each $q \in [Q]$ we associate a \textit{non-zero value }$\lambda_q>0$ and a \textit{location} in the core tensor, denoted by the multi-index $\kq \equiv (\textrm{k}_{q,1},\dots,\textrm{k}_{q,M})$ where $\kqm \in [K_m]$. Each $q$ allocates its value $\lambda_q$ to location $\kq$ such that the value of the core tensor at an arbitrary location $\lat \equiv (\textrm{k}_1,\dots, \textrm{k}_M)$ is
\begin{align}
\label{eq:core}
\lambda_{\lat} &= \sum_{q=1}^Q \mathds{1}(\kq=\lat) \, \lambda_q
\end{align}
The name $\allocore$ is a contraction of ``\underline{allo}cated \underline{$\ell_0$}-\underline{co}nstrained \underline{core}'' that is meant to convey this idea.

The \allocore~decomposition can thus be written as
\begin{align}
\label{eq:allocore}
\widehat{\Ytensor} &= \sum_{q=1}^Q \lambda_q \left(\boldsymbol{\phi}^{\mathsmaller{(1)}}_{\textrm{k}_{q,1}} \circ \dots \circ \boldsymbol{\phi}^{\mathsmaller{(M)}}_{\textrm{k}_{q,M}}\right)
\end{align}
Like Tucker and unlike CP, this decomposition posits a different latent dimensionality $K_m$ for every mode, and can represent any of the $\prod_{m=1}^M K_m$ possible latent classes. However, unlike Tucker and like CP, the number of represented classes is capped at a small number $Q \ll \prod_{m=1}^M K_m$ and thus the parameter space does not suffer Tucker's ``exponential blowup'' as we increase the number of modes $M$ or the latent dimensionalities $K_m$.~\looseness=-1

While~\cref{eq:allocore}~can clearly be computed without ``exponential blowup'' in cost, inference of the (unknown) non-zero locations $\kq$ may still be exponential, since each multi-index can take any of $\prod_{m=1}^M K_m$ values. However, in what follows, we introduce additional probabilistic structure that makes the model amenable to a sampling-based inference scheme which both preserves the sparsity in instantiations of $\boldsymbol{\Lambda}$ and explores new values of $\kq$ in $\mathcal{O}(Q\sum_{m=1}^M K_m)$ time.

\textbf{Latent allocation.} The key motif that facilitates efficient parameter inference in \allocore~is the explicit treatment of the latent non-zero locations $\kq$ as conditionally independent categorical random variables. A very general form for their prior is given by
\begin{align}
\label{eq:categorical}
\kq & \stackrel{\textrm{iid}}{\sim} \textrm{Categorical}(\boldsymbol{\Pi}) &\textrm{for } q \in [Q]
\end{align}
where $\boldsymbol{\Pi}$ is some prior probability tensor of the same shape as $\boldsymbol{\Lambda}$ that sums to 1 over all cells $\sum_{\lat} \pi_{\lat}=1$. \Cref{eq:categorical} simply states that the prior probability of the $q^{\textrm{th}}$ unknown location being $\kq$ is $P(\kq \mid \boldsymbol{\Pi}) \!=\! \pi_{\kq}$.

Although $\boldsymbol{\Pi}$ may seem to suffer the same ``exponential blowup'' that \allocore~seeks to alleviate, we can restrict it to have low-rank structure---e.g.,
\begin{align}
\label{eq:pi}
\boldsymbol{\Pi} = \boldsymbol{\pi}^{\mathsmaller{(1)}} \circ \dots \circ \boldsymbol{\pi}^{\mathsmaller{(M)}}
\end{align}
where $\boldsymbol{\pi}^{\mathsmaller{(m)}} \in \Delta^{K_m\!-\!1}$ is a simplex vector. This corresponds to a rank-1 CP decomposition of $\boldsymbol{\Pi}$, such that $P(\kq \mid \boldsymbol{\Pi}) \!=\! \prod_{m=1}^M \pi^{\mathsmaller{(m)}}_{\kqm}$, and thus avoids any exponential dependence on the number of modes. We note that more structured ways of modeling $\boldsymbol{\Pi}$ are compatible with \allocore~but leave their development for the future and adopt this one in all that follows.~\looseness=-1 

With this prior, inference is then based on re-sampling each index $\kqm \!\in [K_m]$ from its \textit{complete conditional}
\begin{align}
&\nonumber\textrm{for } q \in [Q] \textrm{ and } m \in [M]:\\
\label{eq:allocation}
&\hspace{3em} (\kqm \mid -) \sim P(\kqm  \mid \lat_{q, \neg m}, \Ytensor, -)
\end{align}
which is its distribution conditional on data $\Ytensor$, the other indices in the multi-index, denoted here as $\lat_{q,\neg m}$, and all other latent variables and parameters in the model, represented by the ($-$). We could also simply use $P(\kqm \!=\! k \mid -)$ to refer to this conditional. However, we explicitly include $\Ytensor$ in~\cref{eq:allocation} to highlight that this is a posterior distribution, and explicitly include $\lat_{q, \neg m}$ to emphasize that we re-sample each \textit{sub-index} of $\kq$ while holding all other sub-indices fixed.~\looseness=-1

For any likelihood, the complete conditional in~\cref{eq:allocation} takes the following general form
\begin{align*}
P(\kqm \!=\! k \mid -) &= \frac{\pi_k^{\mathsmaller{(m)}} \, P\big(\Ytensor \mid \widehat{\Ytensor}(\kqm \!=\! k)\big)}{\sum_{k'=1}^{K_m}\pi_{k'}^{\mathsmaller{(m)}} \, P\Big(\Ytensor \mid \widehat{\Ytensor}(\kqm \!=\! k')\Big)}
\end{align*}
where the notation $\widehat{\Ytensor}(\kqm \!=\! k)$ denotes the function in~\cref{eq:allocore} computed after setting $\kqm=k$. 

This probability involves only $K_m$ summands in the denominator. One loop through~\cref{eq:allocation} thus takes $\mathcal{O}(Q\sum_{m=1}^M K_m)$ time for a given likelihood and shape of the observed tensor. By comparison, a na\"ive approach that re-samples each multi-index jointly from $P(\kq \mid -)$ would involve a sum over all core entries, and thus cost $\mathcal{O}(Q\prod_{m=1}^M K_m)$. 

Repeatedly re-sampling the sub-indices from their complete conditionals constitutes a Gibbs sampler---i.e., a Markov chain whose stationary distribution is the exact posterior $P(\lat_1,\dots \lat_Q \mid \Ytensor)$. This can be incorporated into a variety of inference schemes for the other latent variables, $\lambda_q$ and $\Phi^{\mathsmaller{(m)}}$, such as Monte Carlo EM, stochastic variational inference, or full MCMC, the details of which are model-dependent.~\looseness=-1 

We emphasize here that regardless of the approach taken for other variables, the computational benefit of \allocore~is tied to a sampling-based approach for $\kq$. By sampling $\kq$ instead of computing expectations as one would do in EM or variational inference, we maintain a sparse instantiation of the core tensor which allows~\cref{eq:allocore} to be computed with only $Q$ summands. Moreover, by adopting a sampling approach, \allocore~explores the state-space of $\kq$ without ``exponential blowup'' in time or memory complexity.

\section{BAYESIAN POISSON \titleallocore}
\label{sec:bpallocore}
We now provide a complete example of an~\allocore-based model and corresponding inference algorithm. The model we develop here is tailored to large sparse count tensors and motivated by applications to the analysis of dynamic multilayer networks.

This model assumes each entry of a count tensor $\Ytensor \in \mathbb{N}_0^{D_1 \times \dots \times D_M}$ is conditionally Poisson distributed,
\begin{align}
\label{gen:BPA}
\yentry &\stackrel{\textrm{ind.}}{\sim} \textrm{Pois}\big(\widehat{\textrm{y}}_{\obs}\big),\hspace{1em} \widehat{\textrm{y}}_{\obs} \equiv \sum_{q=1}^Q \lambda_q \prod_{m=1}^M \phimdk{m}{\im}{\kqm}
\end{align}
where the form of $\widehat{\textrm{y}}_{\obs}$ unpacks~\cref{eq:allocore} for a specific $\obs$. 

Following~\cref{eq:categorical,eq:pi}, we assume a rank-1 decomposition of the prior probability tensor $\boldsymbol{\Pi}$, so that the $M$ coordinates of the $Q$ non-zero locations in the core are conditionally independent:
\begin{align}
\kqm &\stackrel{\textrm{ind.}}{\sim} \textrm{Categorical}(\boldsymbol{\pi}^{\mathsmaller{(m)}})
\intertext{We then posit conjugate Dirichlet priors,}
\boldsymbol{\pi}^{\mathsmaller{(m)}} &\stackrel{\textrm{ind.}}{\sim} \textrm{Dirichlet}(\tfrac{\alpha^{(m)}_0}{D_m}, \dots, \tfrac{\alpha^{(m)}_0}{D_m}),\label{eq:priorpi}
\intertext{where $\alpha^{(m)}_0$ is a hyperparameter. For both the non-zero core values and the entries of the factor matrices, we posit independent gamma priors, which are conditionally conjugate to the Poisson likelihood:}
\label{eq:priorlam}
\lambda_q &\stackrel{\textrm{iid}}{\sim}\textrm{Gam}(a_0,\, b_0), \\
\label{eq:priorphi}
\phi_{d,k}^{\mathsmaller{(m)}} &\stackrel{\textrm{iid}}{\sim}\textrm{Gam}(e_0,\, f_0) \hspace{1em} d \in [D_m], k \in [K_m]
\end{align}

\textbf{Full MCMC inference.} With a standard auxiliary variable scheme, this model is amenable to full Bayesian inference via Gibbs sampling for all parameters. It is also possible to adopt a hybrid approach that interleaves Gibbs sampling for the non-zero core locations (\cref{eq:allocation}) with point estimation or variational inference of the model parameters; we leave exploration of such schemes for future work.

As with all ``allocative'' Poisson models~\citep{schein2019allocative,yildirim2020bayesian}, the likelihood in~\cref{gen:BPA} can be equivalently expressed so that the observed count $\yentry = \sum_{q=1}^Q \textrm{y}_{\obs,q}$ is assumed equal to a sum of ``latent sources''~\citep{cemgil_bayesian_2009}, each of which is a conditionally independent Poisson random variable
\begin{align}
\label{eq:sources}
\textrm{y}_{\obs,q} \stackrel{\textrm{ind.}}\sim \Pois{\widehat{\textrm{y}}_{\obs,q}},\hspace{1em}
\widehat{\textrm{y}}_{\obs,q} &\equiv \lambda_q \prod_{m=1}^M \phimdk{m}{\im}{\kqm}
\end{align}
While the gamma priors in~\cref{eq:priorlam,eq:priorphi} are not directly conjugate to the likelihood of $\yentry$ due to the sum in its rate, they are conjugate to~\cref{eq:sources}. An MCMC approach therefore begins by sampling the latent sources as auxiliary variables from their complete conditional, which is multinomial:
\begin{align*}
\left(\big(\textrm{y}_{\obs,q}\big)_{q=1}^Q \mid -\right) &\sim \textrm{Multi}\left(\yentry,\, \big(\tfrac{\widehat{\textrm{y}}_{\obs,q}}{\sum_{q'=1}^Q \widehat{\textrm{y}}_{\obs,q'}}\big)_{q=1}^Q\right)
\end{align*}
This is commonly referred to as the ``thinning step'' in the Poisson factorization literature and represents the main computational bottleneck. Notice that it need only be run for $\yentry > 0$ and thus scales well with only the observed non-zeros $||\Ytensor||_0$. Bayesian Poisson Tucker decomposition~\citep{schein_bayesian_2016} features an analogous step---however, there, $\yentry$ is thinned across all $\prod_{m=1}^M K_m$ latent classes of the Tucker decomposition, and thus scales as $\mathcal{O}(||\Ytensor||_0 \cdot |\boldsymbol{\Lambda}|)$, whereas here the thinning step is only $\mathcal{O}(||\Ytensor||_0 \cdot ||\boldsymbol{\Lambda}||_0)$.

Conditioned on values of all latent sources $\textrm{y}_{\obs,q}$, the complete conditionals for all other parameters are available in closed form via conjugacy and can be sampled from efficiently---we relegate the statement and derivation of their specific forms to the appendix.

\section{RELATED WORK}
\label{sec:connections}

The basic idea of promoting sparse representations in tensor or matrix decompositions has been widespread in the literature for decades. The majority of such work introduces \textit{soft} sparsity constraints that encourage sparsity but do not guarantee any specific degree of it. This is most commonly achieved by penalizing the $\ell_1$-norm of model parameters, which can be incorporated into loss functions while preserving convexity. Such work goes back to at least the non-negative sparse basis and coding methods of \citet{hoyer_non-negative_2002,hoyer_non-negative_2004}. Penalizing the $\ell_0$-norm of model parameters introduces a non-convex constraint, and much less work has embraced hard sparsity constraints as a result.~\citet{morup_approximate_2008}, for instance, motivate $\ell_0$ constraints, but then approximate them with convex $\ell_1$ constraints.

Much of the existing work on $\ell_0$-constrained decompositions are for non-negative \textit{matrix} factorization~\citep{peharz_sparse_2012,bolte_proximal_2014,cohen_nonnegative_2019,de_leeuw_novel_2020,wu_co-sparse_2022,nadisic_matrix-wise_2022}. An exception is that of~\citet{kiers_candecompparafac_2020} who develop an $\ell_0$-constrained CP decomposition method, which consists of truncating small values in the factor matrices to zero during inference.

A number of papers present approaches involving the phrase
``sparse Tucker''. This typically refers to sparsity in the observed tensor, not the core. However, a few recent papers introduce sparsity to the core tensor~\citep{park_vest_2021,fang_bayesian_2021,zhang_sparse_2022}. While these approaches introduce interesting new motifs, such as spike-and-slab priors, they do not exploit that sparsity to improve the computational efficiency of Tucker, with some having quadratic or even cubic dependence on the size of the core tensor.~\looseness=-1

There is a rich tradition of CP decomposition models for large sparse count tensors or contingency tables under Poisson or multinomial likelihoods~\citep[inter alia]{welling_positive_2001,dunson_nonparametric_2009,chi_tensors_2012,zhou_bayesian_2015,schein_bayesian_2015}.~\looseness=-1

There is much less work on Tucker decomposition for count tensors.~\citet{bhattacharya_simplex_2012} introduced Tucker under a multinomial likelihood and showed its relation to latent class models. \citet{johndrow_tensor_2017} further developed a ``collapsed'' Tucker model which allocates the $M$ sub-indices of each latent class to $M' \leq M$ groups, thus interpolating between CP ($M'\!=\!1$) and full Tucker ($M'\!=\!M$); this work shares many themes with ours, but is distinct.~\looseness=-1

Finally, multiple recent convergent lines of work advocate for the use of Poisson Tucker decomposition in analyzing complex multilayer networks~\citep{schein_bayesian_2016,de_bacco_community_2017,aguiar_tensor_2023,stoehr_ordered_2023}, as its structure can be seen to unify a number of community block models in the statistical networks literature. While it can be applied more generally,~\allocore~draws inspiration from this line of work and is intended to allow Tucker's promise to be practically achievable in the analysis of large networks.~\looseness=-1

\section{COMPLEX NETWORK DATA}
\label{sec:realdata}

In this section we apply the \allocore~model in~\cref{sec:bpallocore} to two real datasets of dynamic multilayer networks of nation-state actors. We evaluate~\allocore's predictive performance as a function of wall-clock time as we vary $Q$, and compare it to CP and full Tucker decomposition, using both our own implementation of Gibbs sampling, as well as a recent state-of-the-art implementation of EM~\citep{aguiar_tensor_2023}. We also provide an illustrative exploration of the interpretable latent structure inferred by~\allocore.

\subsection{International Relations Event Datasets}
Data of the form ``country $i$ took action $a$ to country $j$ at time step $t$'' are routinely analyzed in political science~\citep{schrodt_event_1995}. Such data can be viewed as a dynamic multilayer network of $V$ country actors with $A$ layers (i.e., action types) that evolves over $T$ time steps, and can be represented as a 4-mode count tensor $\Ytensor \in \mathbb{N}_0^{V \times V \times A \times T}$, where each entry $\textrm{y}_{i,j,a,t}$ records the number of times $i$ took action $a$ to $j$ during time step $t$. We adopt the notation $\textrm{y}_{i,j,a,t} \equiv \textrm{y}_{i \xrightarrow{a} j}^{\mathsmaller{(t)}}$. 

A line of papers advocate for Tucker decompositions to analyze international relations event data~\citep{hoff_multilinear_2015,hoff_equivariant_2016,schein_bayesian_2016,minhas_new_2016,stoehr_ordered_2023}, and we follow in this line by analyzing two of such datasets with \allocore. The 4-mode core tensor $\boldsymbol{\Lambda}$, in this setting, can be interpreted as capturing community--community interactions, where an entry $\lambda_{c,d,k,r}$ is the rate at which the $c^{\textrm{th}}$ \textit{community} of sender countries takes actions in \textit{topic} $k$ towards the $d^{\textrm{th}}$ \textit{community} of receiver countries when in \textit{temporal regime} $r$. We adopt the notation $\lambda_{c,d,k,r} \equiv \lambda_{c \xrightarrow{k} d}^{\mathsmaller{(r)}}$.

\textbf{TERRIER.} We construct the first tensor using the Temporally Extended, Regular, Reproducible International Event Records (TERRIER) dataset, recently released by~\cite{halterman_adaptive_2017}. This dataset codes for $V\!=\!206$ country actors and adopts the CAMEO coding scheme for action types~\citep{gerner_conflict_2002}, wherein $A=20$ high-level action types range in severity from $\textsc{Make Public Statement}$ ($a=1$) to $\textsc{Engage In Mass Violence}$ ($a=20$). The observation window ranges from January 2000 to December 2006, which we bin monthly to obtain $T=84$ time steps. The resulting tensor $\Ytensor \in \mathbb{N}_0^{206 \times 206 \times 20 \times 84}$ is 99.5\% sparse.\looseness=-1

\textbf{ICEWS.} We construct the second tensor using the Integrated Crisis Early Warning System (ICEWS) dataset~\citep{boschee_icews_2015}. ICEWS also uses the CAMEO coding scheme ($A=20$) but represents slightly more $V=249$ country actors, and has a longer observation window from 1995 to 2013, which we bin yearly to obtain $T=19$. The resulting count tensor $\Ytensor \in \mathbb{N}_0^{249 \times 249 \times 20 \times 19}$ is 98.5\% sparse.

\subsection{Experimental Design and Settings}

\textbf{Design.} To evaluate predictive performance, we randomly generate multiple train--test splits of the ICEWS and TERRIER tensors. Each split holds out a set of \textit{fibers} of the tensor---i.e., vectors indexed by coordinates to all modes but one. In this case, we select a set of time fibers---i.e., we randomly select combinations of sender, receiver, and action type $(i,j,a)$ and hold out the vector $\textbf{y}_{i \xrightarrow{a} j} \in \mathbb{N}_0^T$ for each one. In each split, we hold out a random 1\% of all $V\cdot V \cdot A$ such fibers, and generate 8 such splits for each tensor.

\textbf{AL$\ell_0$CORE~settings.} We implemented the Gibbs sampling algorithm for the \allocore~model in~\cref{sec:bpallocore} using the programming language Julia. In all experiments, we discard the first 1,000 samples as burn-in and then run MCMC for another 4,000 iterations, saving every $20^{\textrm{th}}$ sample. This returns a set of $S=200$ posterior samples of the model parameters, which can be used to compute a set of reconstructions $\big(\widehat{\Ytensor}^{(s)}\big)_{s=1}^S$.

Throughout these experiments, we vary $Q$ to see its effect on both predictive performance and run-time. Since the computational cost is determined by $Q$ (rather than the size of the core), we always set $K_1=\dots=K_M=Q$, for simplicity, and further initialize the core tensor to be super-diagonal $\boldsymbol{\Lambda}=\textrm{diag}(\lambda_1,\dots,\lambda_Q)$. We refer to this family as \textit{canonical \allocore}, since it contains CP as a special case. We further set the hyperparameters to standard default values $a_0{=}b_0{=}e_0{=}1.0$, $f_0{=}10$, and $\alpha_0^{\mathsmaller{(m)}}{=}0.1$.~\looseness=-1

\textbf{Baselines.} We compare \allocore~to three baselines: 1) CP using Gibbs sampling, 2) full Tucker using Gibbs sampling, and 3) full Tucker using EM. All three models adopt a Poisson likelihood, as given in the left-hand side of~\cref{gen:BPA}. The Gibbs sampling baselines further adopt analogous prior distributions. We implement them in Julia using much of the same code as for \allocore, and run them using the same settings. For the third baseline, we use a recent state-of-the-art implementation of EM for 3-mode Tucker known as NNTuck~\citep{aguiar_tensor_2023}, and  adapt its Python code for 4-mode Tucker~\looseness=-1.

\begin{figure}[t]
    \centering
    \includegraphics[width=\linewidth]{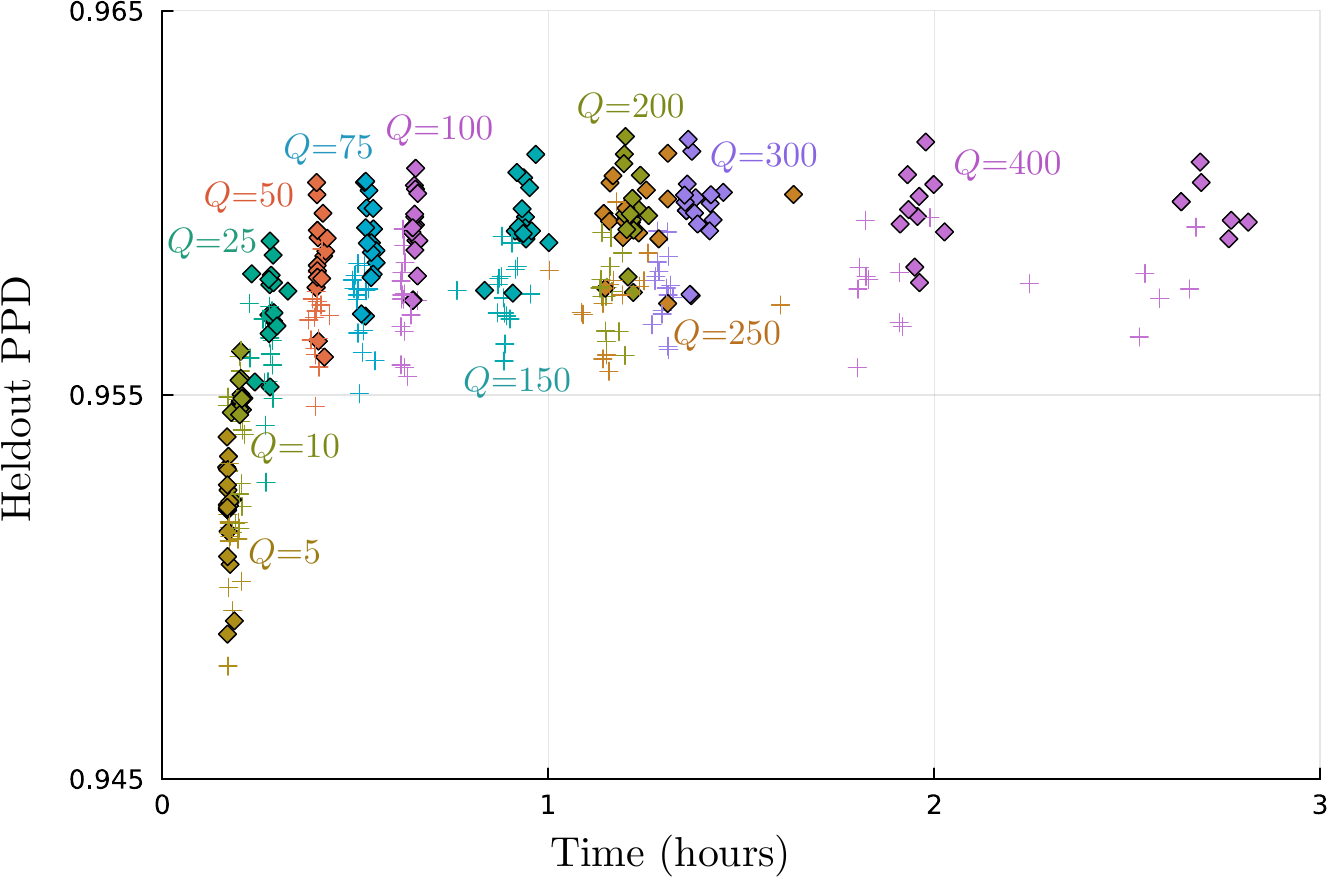}
    \caption{PPD by wall-clock time. Each point is an~\allocore~model run for 5,000 iterations, with a certain $Q$, denoted by color, and either the large or small core shape, denoted by $\diamond$ or $+$, respectively. Performance plateaus early, suggesting $\allocore$~can match full Tucker at a small fraction of cost.}
    \label{fig:DenseTuckerUnnecessary}~\looseness=-1
\end{figure} 
\textbf{Evaluation criteria.}  For each tensor and train-test split, we fit each model to the training set, and evaluate the pointwise predictive density (PPD) it assigns to heldout data---this is defined as
\begin{align*}
        \textrm{PPD}(\mathcal{H}) = \exp\Big(\frac{1}{|\mathcal{H}|}\sum_{\mathbf{d} \in \mathcal{H}} \log \Big[\frac{1}{S}\sum_{s=1}^S \textrm{Pois}(\yentry; \widehat{\textrm{y}}_{\obs}^{(s)})\Big]\Big),
\end{align*}
where $\mathcal{H}$ is the set of multi-indices in the test set. We also examine PPD on the positive-only part of the test set---i.e., $\textrm{PPD}(\mathcal{H}_{>0})$ where $\mathcal{H}_{>0} \equiv \{\obs \in \mathcal{H}: \yentry >0\}$. 

        \begin{figure}[!t]
        \centering
            \includegraphics[width=\linewidth]{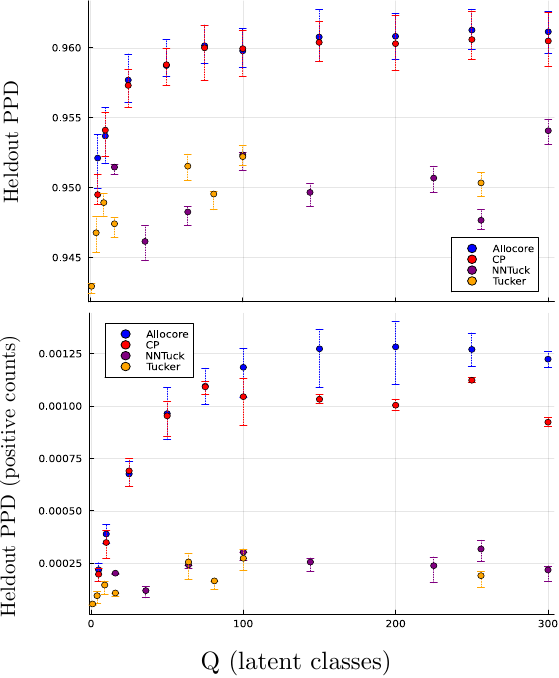}
        \caption{ PPD across models on the full heldout set (top) and the positive-only heldout (bottom) across $Q$. Error bars span the interquartile range across masks.}
        \label{fig:ppd}
    \end{figure}

\subsection{Predictive Results}
We first examine the predictive performance of~\allocore~as $Q$ increases. Note that as $Q \rightarrow |\boldsymbol{\Lambda}|$, $\allocore$ becomes full Tucker decomposition. We can thus ask: at what fraction of core density $\nicefrac{Q}{|\boldsymbol{\Lambda}|}$ (and thus fraction of computational cost), does \allocore~achieve all the benefits of Tucker?

We consider both large and small core shapes. The small core shape we use is the one used by~\citet{schein_bayesian_2016} for the same data: $20 \times 20 \times 6 \times 3$. The large core shape is $50 \times 50 \times 6 \times 10$. For each core shape, we then consider values of $Q$ ranging from 5 to 400. Note that $Q=400$ still represents only a small fraction of full Tucker for both the small (5.56\% dense) and large (0.27\% dense) cores.~\looseness=-1

\Cref{fig:DenseTuckerUnnecessary} plots heldout PPD as a function of wall-clock time for all unique combinations of budget $Q$, core shape, and train-test splits of TERRIER. Performance plateaus very early for both core shapes, around $Q=150$, which takes approximately $1$ hour for 5,000 MCMC iterations. By comparison, one iteration of EM for full Tucker (NNTuck), using the small core shape, takes around 125 seconds. \citet{aguiar_tensor_2023}~recommend running NNTuck for 1,000 iterations, which would take around 1.5 days for a single run (they further recommend 20 random restarts). It is worth emphasizing that NNTuck represents the state-of-the-art for Poisson Tucker decomposition and that its severe run-time in this setting reflects the fundamental problem with all Tucker decomposition.

We further compare \allocore~to all of the CP and Tucker decomposition baselines. To fairly compare across these models, we set $Q$ in CP and \allocore~to be equal and further set the shape of the core tensor for the full Tucker methods (Gibbs sampling and NNTuck) such that the core has roughly $Q$ total entries. We consider $Q \in \{5, 10, 25, 50, 75, 100, 150, 200, 250, 300\}$ and fit every model with every $Q$ to every train-test split.

Figure~\ref{fig:ppd} reports the results. \allocore~performs substantially better than either Tucker approach. Moreover, as $Q$ increases, CP's performance degrades, suggesting possible overfitting, while $\allocore$ appears robust. We note this tells only part of the story, as the full Tucker approaches took substantially longer to run, even while performing substantially worse.~\looseness=-1

We report in the appendix a number of other predictive results, similar to those reported here, as well as a toy synthetic data experiment seeking to better understand \allocore's shrinkage properties.

\subsection{Qualitative Results}
\label{sec:qual}

Finally, we fit \allocore~to the fully-observed ICEWS dataset and explore its inferred latent structure. For this, we binned time steps binned by month (rather than year) to explore more fine-grained temporal structure, which yields $T=228$ time steps, and a resulting count tensor $\Ytensor$ of shape $249 \times 249 \times 20 \times 228$.

In this setting, the core tensor has shape $C \times D \times K \times R$. Following~\citet{schein_bayesian_2016}, we interpret $C$ and $D$ to be the number of \textit{communities} of sender and receiver countries, respectively, $K$ to be the number of \textit{action topics}, and $R$ to be the number of \textit{temporal regimes}. We set these to $C \!=\! D \!=\!50$, $K \!=\! 20$, and $R \!=\! 300$. The number of possible latent classes is $C^2DK\!=$15,000,000. By comparison~\citet{schein_bayesian_2016} used a $20 \times 20 \times 6 \times 3$ core, resulting in 7,200, while~\citet{schein_bayesian_2015} used a CP decomposition with only 50 latent classes. \allocore~is able to explore such a large latent space due to its sparsity constraint, which we set here to $Q=400$.

We ran 10,000 MCMC iterations and used the last posterior sample for exploratory analysis. We first inspected how many classes were inferred to be non-zero. Recall that $Q$ is merely an upper bound on the non-zero classes. We find that only 215 (of the possible 400) core elements are non-zero. Moreover, we find that the inferred non-zero values $\lambda_{\lat}$ for those 215 classes are highly unequal, with only 44 classes accounting for 50\% of the total mass i.e., $\sum_{\lat} \lambda_\lat$, and 150 accounting for 95\%. It is intriguing that \allocore~shrinks to this particular range of 50--150 classes, which is the range typically used to apply CP to the same data.~\looseness=-1

\begin{figure*}[h!]
    \centering
    \includegraphics[width=\textwidth]{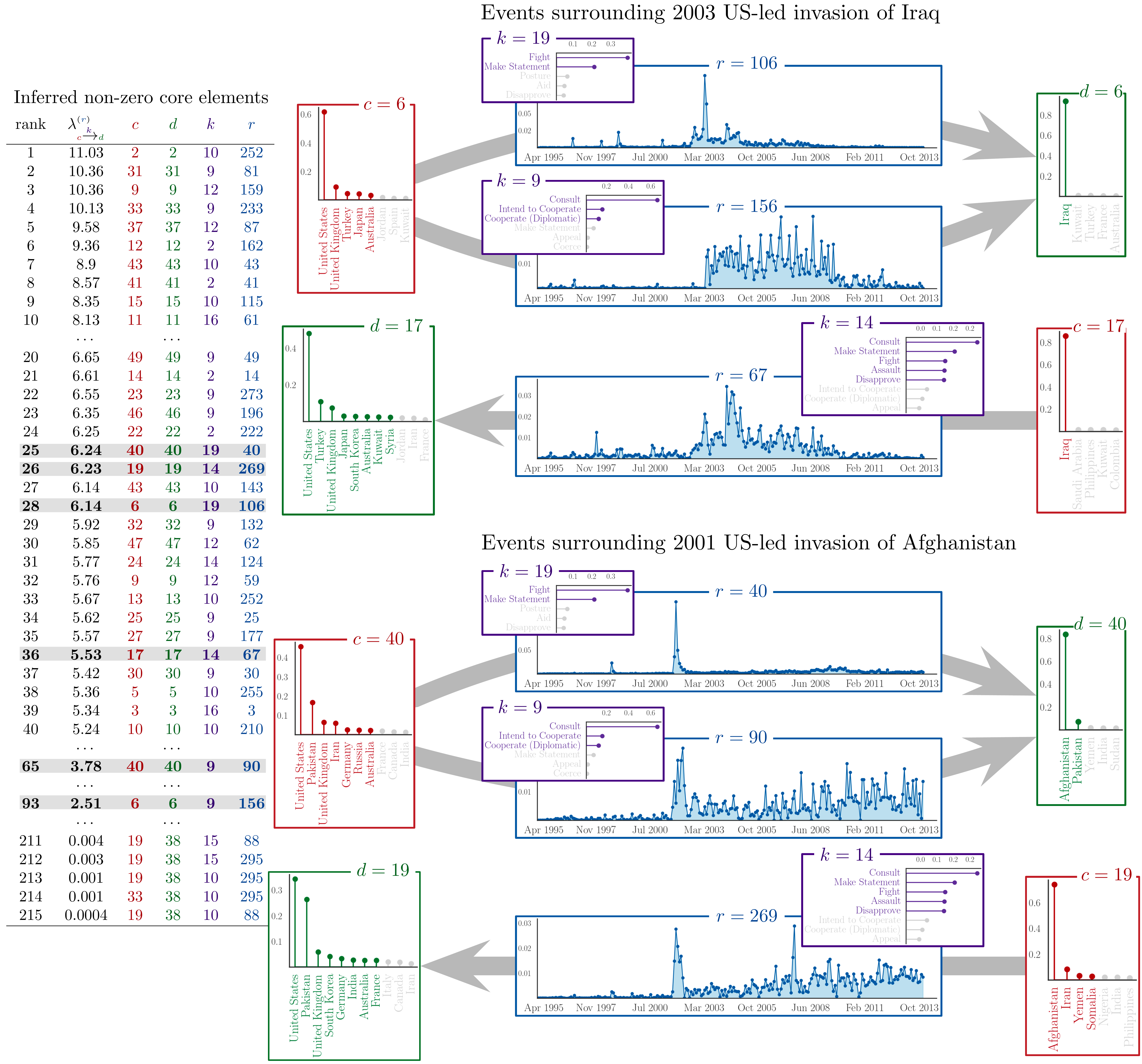}
    \caption{Example of the latent class structure inferred by \allocore~with $Q=400$ on the ICEWS dataset.\\ \textit{Left:} The inferred non-zero locations in the core tensor, with rows sorted according to the inferred values $\lambda_{c \xrightarrow{k} d}^{\mathsmaller{r}}$. \textit{Right:} Two sets of three inferred classes, each corresponding to a major war, and each following the same cadence---i.e., $c \xrightarrow{19} d'$, $c \xrightarrow{9} d'$, $c' \xrightarrow{14} d$---described in~\cref{sec:qual}. The blue stem plots depict all elements in a given time-step factor, while red, green, and purple stem plots depict only the largest elements in a given sender, receiver, and action factor, respectively; in these, stems are greyed out when normalized values fall below 0.02.}
    \label{fig:map}
\end{figure*}

We provide visualizations of the top 100 inferred classes, ranked by their value $\lambda_{\lat}$, in~\cref{sec:app_qualitative} and a more detailed visualization of six of those classes in~\cref{fig:map}. Each class in this setting corresponds to a location $(c,d,k,r)$ in the core  for which the inferred value $\lambda_{c,d,k,r} \equiv \lambda_{c \xrightarrow{k} d}^{\mathsmaller{(r)}}$ is non-zero. We can interpret a given class by inspecting the four factor vectors it indexes---i.e., the $c^{\textrm{th}}$ and $d^{\textrm{th}}$ community factors, $\boldsymbol{\phi}^{\mathsmaller{(1)}}_c \in \mathbf{R}_+^V$ and $\boldsymbol{\phi}^{\mathsmaller{(2)}}_d \in \mathbf{R}_+^V$, over sender and receiver countries, respectively, the $k^{\textrm{th}}$ topic factor $\boldsymbol{\phi}^{\mathsmaller{(3)}}_k \in \mathbf{R}_+^A$ over actions, and the $r^{\textrm{th}}$ regime factor $\boldsymbol{\phi}^{\mathsmaller{(4)}}_r \in \mathbf{R}_+^T$ over time steps. Since these are non-negative, simply inspecting the largest elements in each factor tends to yield highly intuitive interpretations.

In exploring the inferred classes, we found one particularly interesting pattern. For multiple major wars that occurred during the observation window 1995--2013, we find two pairs of sender and receiver communities, $(c,d)$ and $(c',d')$, which correspond to the two sets of belligerents on either side of the war. Furthermore, for all these wars, we find a shared cadence involving three inferred classes. In the first class, community $c$ takes actions in topic $k=19$ to community $d'$---i.e., $c \xrightarrow{19} d'$---for some regime $r$. This topic $k=19$, places most of its weight on the action \textsc{Fight}. In the second class, the same $c$ takes actions in topic $k=9$ to the same $d'$---i.e., $c \xrightarrow{9} d'$---with this topic placing its weight on cooperative actions, such as \textsc{Intend to Cooperate}. In the third class, sender community $c'$ (which corresponds roughly to the same belligerents in receiver community $d'$) takes actions in topic $k=14$ towards community $d$---i.e., $c' \xrightarrow{14} d$. We find this same cadence of three classes---i.e., $c \xrightarrow{19} d'$, $c \xrightarrow{9} d'$, $c' \xrightarrow{14} d$---for five pairs of $(c,d)$ and $(c',d')$, which correspond to belligerents on either side of 1) the 2001 US-led invasion of Afghanistan, 2) the 2003 US-led invasion of Iraq, 3) the 2006 Israel--Hezbollah war, 4) the Israeli-Palestinian conflict, and 5) the Yugoslav wars. We visualize this cadence for two of these wars in figure~\ref{fig:map}. We emphasize that surfacing this kind of pattern is facilitated by \allocore's ability to share factors across classes (something CP cannot do) but to do so \textit{selectively} without suffering Tucker's ``exponential blowup''.~\looseness=-1

\section{CONCLUSION} 
The Tucker decomposition's qualitative appeal has always been difficult to practically achieve due to the ``exponential blowup'' associated with increasing the number of latent factors. For many problems involving large sparse count tensors, such as the analysis of complex multilayer networks, where modelers often seek to represent many communities and their interactions, the disparity between Tucker's conceptual appeal and its practicality is stark.  The~\allocore~decomposition remedies this disparity by leveraging sparsity to decouple the size of the core tensor from the computational cost of inference. Although particularly motivated for the analysis of sparse count data, the motif of allocated sparsity could be incorporated into many other models and methods, development of which are promising avenues of future work.

\ackaccepted

This material is based upon work supported by the National Science Foundation Graduate Research Fellowship Program under Grant No. 2140001.

\bibliography{bibliography}

\begin{thebibliography}{}

\bibitem[Aguiar et~al., 2023]{aguiar_tensor_2023}
Aguiar, I., Taylor, D., and Ugander, J. (2023).
\newblock A tensor factorization model of multilayer network interdependence.

\bibitem[Amjad et~al., 2019]{amjad_mrsc_2019}
Amjad, M., Misra, V., Shah, D., and Shen, D. (2019).
\newblock {mRSC}: {Multi}-dimensional {Robust} {Synthetic} {Control}.
\newblock {\em Proceedings of the ACM on Measurement and Analysis of Computing Systems}, 3(2):1--27.

\bibitem[Bhattacharya and Dunson, 2012]{bhattacharya_simplex_2012}
Bhattacharya, A. and Dunson, D.~B. (2012).
\newblock Simplex {Factor} {Models} for {Multivariate} {Unordered} {Categorical} {Data}.
\newblock {\em Journal of the American Statistical Association}, 107(497):362--377.

\bibitem[Bolte et~al., 2014]{bolte_proximal_2014}
Bolte, J., Sabach, S., and Teboulle, M. (2014).
\newblock Proximal alternating linearized minimization for nonconvex and nonsmooth problems.
\newblock {\em Mathematical Programming}, 146(1-2):459--494.

\bibitem[Boschee et~al., 2015]{boschee_icews_2015}
Boschee, E., Lautenschlager, J., O'Brien, S., Shellman, S., Starz, J., and Ward, M. (2015).
\newblock {ICEWS} {Coded} {Event} {Data}.

\bibitem[Cemgil, 2009]{cemgil_bayesian_2009}
Cemgil, A.~T. (2009).
\newblock Bayesian inference for nonnegative matrix factorisation models.
\newblock {\em Computational Intelligence and Neuroscience}, 2009:785152.

\bibitem[Chi and Kolda, 2012]{chi_tensors_2012}
Chi, E.~C. and Kolda, T.~G. (2012).
\newblock On {Tensors}, {Sparsity}, and {Nonnegative} {Factorizations}.
\newblock {\em SIAM Journal on Matrix Analysis and Applications}, 33(4):1272--1299.

\bibitem[Cichocki et~al., 2015]{cichocki_tensor_2015}
Cichocki, A., Mandic, D., De~Lathauwer, L., Zhou, G., Zhao, Q., Caiafa, C., and Phan, H.~A. (2015).
\newblock Tensor {Decompositions} for {Signal} {Processing} {Applications}: {From} two-way to multiway component analysis.
\newblock {\em IEEE Signal Processing Magazine}, 32(2):145--163.

\bibitem[Cohen and Gillis, 2019]{cohen_nonnegative_2019}
Cohen, J.~E. and Gillis, N. (2019).
\newblock Nonnegative {Low}-rank {Sparse} {Component} {Analysis}.
\newblock In {\em 2019 {IEEE} {International} {Conference} on {Acoustics}, {Speech} and {Signal} {Processing}}, pages 8226--8230.

\bibitem[De~Bacco et~al., 2017]{de_bacco_community_2017}
De~Bacco, C., Power, E.~A., Larremore, D.~B., and Moore, C. (2017).
\newblock Community detection, link prediction, and layer interdependence in multilayer networks.
\newblock {\em Physical Review E}, 95(4):042317.

\bibitem[de~Leeuw, 2020]{de_leeuw_novel_2020}
de~Leeuw, D. (2020).
\newblock {\em A Novel $\ell_0$-Sparse {Nonnegative} {Matrix} {Factorization} {Algorithm} ($\ell_0$-{SNMF}) for {Community} {Detection}. {An} {Approach} {Using} {Iterative} {Majorization}}.
\newblock PhD thesis, Erasmus University Rotterdam, School of Economics.

\bibitem[Dunson and Xing, 2009]{dunson_nonparametric_2009}
Dunson, D.~B. and Xing, C. (2009).
\newblock Nonparametric {Bayes} {Modeling} of {Multivariate} {Categorical} {Data}.
\newblock {\em Journal of the American Statistical Association}, 104(487):1042--1051.

\bibitem[Fang et~al., 2021]{fang_bayesian_2021}
Fang, S., Kirby, R.~M., and Zhe, S. (2021).
\newblock Bayesian streaming sparse {Tucker} decomposition.
\newblock In {\em Proceedings of the {Thirty}-{Seventh} {Conference} on {Uncertainty} in {Artificial} {Intelligence}}, pages 558--567.

\bibitem[Gerner et~al., 2002]{gerner_conflict_2002}
Gerner, D., Jabr, R., and Schrodt, P. (2002).
\newblock Conflict and {Mediation} {Event} {Observations} ({CAMEO}): {A} {New} {Event} {Data} {Framework} for the {Analysis} of {Foreign} {Policy} {Interactions}.
\newblock {\em International Studies Association, New Orleans}.

\bibitem[Halterman et~al., 2017]{halterman_adaptive_2017}
Halterman, A., Irvine, J., Landis, M., Jalla, P., Liang, Y., Grant, C., and Solaimani, M. (2017).
\newblock Adaptive scalable pipelines for political event data generation.
\newblock In {\em 2017 {IEEE} {International} {Conference} on {Big} {Data}}, pages 2879--2883.

\bibitem[Harshman, 1970]{harshman_foundations_1970}
Harshman, R. (1970).
\newblock Foundations of the {PARAFAC} procedure: {Models} and conditions for an ``explanatory'' multi-model factor analysis.
\newblock {\em UCLA Working Papers in Phonetics}, 16:1--84.

\bibitem[Hoff, 2015]{hoff_multilinear_2015}
Hoff, P.~D. (2015).
\newblock Multilinear tensor regression for longitudinal relational data.
\newblock {\em The Annals of Applied Statistics}, 9(3):1169--1193.

\bibitem[Hoff, 2016]{hoff_equivariant_2016}
Hoff, P.~D. (2016).
\newblock Equivariant and {Scale}-{Free} {Tucker} {Decomposition} {Models}.
\newblock {\em Bayesian Analysis}, 11(3).

\bibitem[Hoyer, 2002]{hoyer_non-negative_2002}
Hoyer, P. (2002).
\newblock Non-negative sparse coding.
\newblock {\em Proceedings of the 12th IEEE Workshop on Neural Networks for Signal Processing}, pages 557--565.

\bibitem[Hoyer, 2004]{hoyer_non-negative_2004}
Hoyer, P.~O. (2004).
\newblock Non-negative {Matrix} {Factorization} with {Sparseness} {Constraints}.
\newblock {\em The Journal of Machine Learning Research}, 5:1457--1469.

\bibitem[Johndrow et~al., 2017]{johndrow_tensor_2017}
Johndrow, J.~E., Bhattacharya, A., and Dunson, D.~B. (2017).
\newblock Tensor decompositions and sparse log-linear models.
\newblock {\em The Annals of Statistics}, 45(1):1--38.

\bibitem[Kiers and Giordani, 2020]{kiers_candecompparafac_2020}
Kiers, H. A.~L. and Giordani, P. (2020).
\newblock {CANDECOMP}/{PARAFAC} with zero constraints at arbitrary positions in a loading matrix.
\newblock {\em Chemometrics and Intelligent Laboratory Systems}, 207:104145.

\bibitem[Kolda and Bader, 2009]{kolda_tensor_2009}
Kolda, T.~G. and Bader, B.~W. (2009).
\newblock Tensor {Decompositions} and {Applications}.
\newblock {\em SIAM Review}, 51(3):455--500.

\bibitem[Lee and Seung, 1999]{lee_learning_1999}
Lee, D.~D. and Seung, H.~S. (1999).
\newblock Learning the parts of objects by non-negative matrix factorization.
\newblock {\em Nature}, 401(6755):788--791.

\bibitem[Marot et~al., 2008]{marot2008advances}
Marot, J., Fossati, C., and Bourennane, S. (2008).
\newblock About advances in tensor data denoising methods.
\newblock {\em EURASIP Journal on Advances in Signal Processing}, 2008:1--12.

\bibitem[Minhas et~al., 2016]{minhas_new_2016}
Minhas, S., Hoff, P.~D., and Ward, M.~D. (2016).
\newblock A new approach to analyzing coevolving longitudinal networks in international relations.
\newblock {\em Journal of Peace Research}, 53(3):491--505.

\bibitem[Morup et~al., 2008]{morup_approximate_2008}
Morup, M., Madsen, K.~H., and Hansen, L.~K. (2008).
\newblock Approximate {L$_0$} constrained non-negative matrix and tensor factorization.
\newblock In {\em 2008 {IEEE} {International} {Symposium} on {Circuits} and {Systems}}, pages 1328--1331.

\bibitem[Nadisic et~al., 2022]{nadisic_matrix-wise_2022}
Nadisic, N., Cohen, J., Vandaele, A., and Gillis, N. (2022).
\newblock Matrix-wise $\ell_0$-constrained sparse nonnegative least squares.
\newblock {\em Machine Learning}, 111:1--43.

\bibitem[Park et~al., 2021]{park_vest_2021}
Park, M., Jang, J.-G., and Sael, L. (2021).
\newblock {VEST}: {Very} {Sparse} {Tucker} {Factorization} of {Large}-{Scale} {Tensors}.
\newblock In {\em 2021 {IEEE} {International} {Conference} on {Big} {Data} and {Smart} {Computing}}, pages 172--179.

\bibitem[Peharz and Pernkopf, 2012]{peharz_sparse_2012}
Peharz, R. and Pernkopf, F. (2012).
\newblock Sparse nonnegative matrix factorization with $\ell_0$-constraints.
\newblock {\em Neurocomputing}, 80(1):38--46.

\bibitem[Schein, 2019]{schein2019allocative}
Schein, A. (2019).
\newblock Allocative {P}oisson factorization for computational social science.

\bibitem[Schein et~al., 2015]{schein_bayesian_2015}
Schein, A., Paisley, J., Blei, D.~M., and Wallach, H. (2015).
\newblock Bayesian {Poisson} {Tensor} {Factorization} for {Inferring} {Multilateral} {Relations} from {Sparse} {Dyadic} {Event} {Counts}.
\newblock In {\em Proceedings of the 21th {ACM} {SIGKDD} {International} {Conference} on {Knowledge} {Discovery} and {Data} {Mining}}.

\bibitem[Schein et~al., 2016]{schein_bayesian_2016}
Schein, A., Zhou, M., Blei, D.~M., and Wallach, H. (2016).
\newblock Bayesian {Poisson} {Tucker} decomposition for learning the structure of international relations.
\newblock In {\em Proceedings of the 33rd {International} {Conference} on {Machine} {Learning}}, pages 2810--2819.

\bibitem[Schrodt et~al., 1995]{schrodt_event_1995}
Schrodt, P., Neack, L., Haney, P., and Hey, J. (1995).
\newblock Event data in foreign policy analysis.
\newblock In {\em Foreign {Policy} {Analysis}: {Continuity} and {Change} in {Its} {Second} {Generation}}, pages 145--166. Prentice Hall.

\bibitem[Stoehr et~al., 2023]{stoehr_ordered_2023}
Stoehr, N., Radford, B.~J., Cotterell, R., and Schein, A. (2023).
\newblock The {Ordered} {Matrix} {Dirichlet} for {State}-{Space} {Models}.
\newblock In {\em Proceedings of {The} 26th {International} {Conference} on {Artificial} {Intelligence} and {Statistics}}, pages 1888--1903.

\bibitem[Tomasi and Bro, 2005]{tomasi2005parafac}
Tomasi, G. and Bro, R. (2005).
\newblock {PARAFAC} and missing values.
\newblock {\em Chemometrics and Intelligent Laboratory Systems}, 75(2):163--180.

\bibitem[Tucker, 1966]{tucker_mathematical_1966}
Tucker, L.~R. (1966).
\newblock Some mathematical notes on three-mode factor analysis.
\newblock {\em Psychometrika}, 31(3):279--311.

\bibitem[Welling and Weber, 2001]{welling_positive_2001}
Welling, M. and Weber, M. (2001).
\newblock Positive tensor factorization.
\newblock {\em Pattern Recognition Letters}, 22(12):1255--1261.

\bibitem[Wu et~al., 2022]{wu_co-sparse_2022}
Wu, F., Cai, J., Wen, C., and Tan, H. (2022).
\newblock Co-sparse {Non}-negative {Matrix} {Factorization}.
\newblock {\em Frontiers in Neuroscience}, 15.

\bibitem[Xiong et~al., 2010]{xiong2010temporal}
Xiong, L., Chen, X., Huang, T.-K., Schneider, J., and Carbonell, J.~G. (2010).
\newblock Temporal collaborative filtering with {B}ayesian probabilistic tensor factorization.
\newblock In {\em Proceedings of the 2010 SIAM international conference on data mining}, pages 211--222. SIAM.

\bibitem[Y{\i}ld{\i}r{\i}m et~al., 2020]{yildirim2020bayesian}
Y{\i}ld{\i}r{\i}m, S., Kurutmaz, M.~B., Barsbey, M., {\c{S}}im{\c{s}}ekli, U., and Cemgil, A.~T. (2020).
\newblock Bayesian allocation model: marginal likelihood-based model selection for count tensors.
\newblock {\em IEEE Journal of Selected Topics in Signal Processing}, 15(3):560--573.

\bibitem[Zhang and Ng, 2022]{zhang_sparse_2022}
Zhang, X. and Ng, M.~K. (2022).
\newblock Sparse nonnegative tucker decomposition and completion under noisy observations.

\bibitem[Zhou et~al., 2015]{zhou_bayesian_2015}
Zhou, J., Bhattacharya, A., Herring, A., and Dunson, D. (2015).
\newblock Bayesian factorizations of big sparse tensors.
\newblock {\em Journal of the American Statistical Association}, 110(512):1562--1576.

\end{thebibliography}

\onecolumn
\twocolumn[]{}
\section*{Checklist}

    \begin{enumerate}

 \item For all models and algorithms presented, check if you include:
 \begin{enumerate}
   \item A clear description of the mathematical setting, assumptions, algorithm, and/or model. \textbf{Yes}
   \item An analysis of the properties and complexity (time, space, sample size) of any algorithm. \textbf{Yes}
   \item (Optional) Anonymized source code, with specification of all dependencies, including external libraries. \textbf{No}
 \end{enumerate}

 \item For any theoretical claim, check if you include:
 \begin{enumerate}
   \item Statements of the full set of assumptions of all theoretical results. \textbf{Not Applicable}
   \item Complete proofs of all theoretical results. \textbf{Not Applicable}
   \item Clear explanations of any assumptions. \textbf{Yes}    
 \end{enumerate}

 \item For all figures and tables that present empirical results, check if you include:
 \begin{enumerate}
   \item The code, data, and instructions needed to reproduce the main experimental results (either in the supplemental material or as a URL). \textbf{Include the \allocore~source code.}
   \item All the training details (e.g., data splits, hyperparameters, how they were chosen). \textbf{Yes} combination in section 6.2 and Supplementary Material.
         \item A clear definition of the specific measure or statistics and error bars (e.g., with respect to the random seed after running experiments multiple times). \textbf{Yes}
         \item A description of the computing infrastructure used. (e.g., type of GPUs, internal cluster, or cloud provider). \textbf{Yes}, cloud provider. 
 \end{enumerate}

 \item If you are using existing assets (e.g., code, data, models) or curating/releasing new assets, check if you include:
 \begin{enumerate}
   \item Citations of the creator If your work uses existing assets. \textbf{Yes}
   \item The license information of the assets, if applicable. \textbf{Not Applicable}
   \item New assets either in the supplemental material or as a URL, if applicable. [\textbf{Yes}/No/Not Applicable] Provide \allocore~source code in supplemental material. 
   \item Information about consent from data providers/curators. \textbf{No}
   \item Discussion of sensible content if applicable, e.g., personally identifiable information or offensive content. \textbf{Not Applicable}
 \end{enumerate}

 \item If you used crowdsourcing or conducted research with human subjects, check if you include:
 \begin{enumerate}
   \item The full text of instructions given to participants and screenshots. \textbf{Not Applicable}
   \item Descriptions of potential participant risks, with links to Institutional Review Board (IRB) approvals if applicable. \textbf{Not Applicable}
   \item The estimated hourly wage paid to participants and the total amount spent on participant compensation. \textbf{Not Applicable}
 \end{enumerate}

 \end{enumerate}
 
\onecolumn

\setcounter{section}{0} 
\renewcommand{\thesection}{\Alph{section}}
\section{DERIVATIONS AND CODE}
\subsection{Complete Conditional Derivations}
We detail and derive the complete conditional updates for posterior inference under Bayesian Poisson~\allocore, leveraging Poisson additivity to streamline the following derivations and use $(\propto_x)$ to denote \textit{proportional to in $x$} (equal up to a constant $c_0$, where $c_0$ does not depend on $x$).

\textbf{Poisson additivity.} For $i \in [n]$, if
$x_i \overset{\textrm{ind.}}{\sim} \textrm{Pois}(c\theta_i)$, then marginally, $x_\bullet \equiv \sum_{i=1}^n x_i \sim \textrm{Pois}(c\sum_{i=1}^n \theta_i).$

\textbf{Complete conditional for $\boldsymbol{\phi^{\mathsmaller{(m)}}_{d,k}}$.}
For each $m \in [M]$, $\textrm{k}_m \in [K_m]$, $d_m \in [D_m]$, we leverage the Poisson additivity of the $\textrm{y}_{\obs, q}$ as follows. 
\begin{align}
    \textrm{y}^{\mathsmaller{(m)}}_{{d_m}{\textrm{k}_m}} &= \sum_{q: \textrm{k}_{q,m} = \textrm{k}_m}\sum_{d' \in \mathcal{D}, d'_m = d_m} \textrm{y}_{d'_q}\,;\\
    \textrm{y}^{\mathsmaller{(m)}}_{{d_m}{\textrm{k}_m}} & \sim \textrm{Pois}(\phi^{\mathsmaller{(m)}}_{d_m, \textrm{k}_m} \underbrace{\sum_{q: \textrm{k}_{q,m} = \textrm{k}_m} \lambda_q \sum_{d' \in \mathcal{D}, d'_m = d_m}\prod_{m'\neq m} \phi^{(m')}_{d_{m'}\textrm{k}_{q, m'}}}_{c^{\mathsmaller{(m)}}_{d_m, \textrm{k}_m}})\,,\\
    \textrm{y}^{\mathsmaller{(m)}}_{{d_m}{\textrm{k}_m}} & \sim \textrm{Pois}(\phi^{\mathsmaller{(m)}}_{d_m, \textrm{k}_m} c^{\mathsmaller{(m)}}_{d_m, \textrm{k}_m})\,;
\end{align}
where for fixed $m$, $\left(\textrm{y}_{d_m, \textrm{k}_m}\right)_{\textrm{k}_m \in [K_m], d_m \in [D_m]}$ are mutually independent. \\
In its complete conditional, factor matrix entry $\phi^{\mathsmaller{(m)}}_{d,k}$ is Gamma-distributed:
\begin{align}
    P(\phi^{\mathsmaller{(m)}}_{d,k} \mid - ) &= P(\phi^{\mathsmaller{(m)}}_{d,k} \mid (\Phi^{(m')})_{m'\neq m}, (\lambda_q, (\textrm{y}_{\obs, q}), (\textrm{k}_{q,m'})_{m'=1}^M)_{q=1}^Q)\\
    &= P(\phi^{\mathsmaller{(m)}}_{d,k} \mid \textrm{y}^{\mathsmaller{(m)}}_{d,k}, c^{\mathsmaller{(m)}}_{d,k})\\
    &\propto_{\phi^{\mathsmaller{(m)}}_{d,k}} P(\textrm{y}^{\mathsmaller{(m)}}_{d,k} \mid  c^{\mathsmaller{(m)}}_{d,k}, \phi^{\mathsmaller{(m)}}_{d,k})  P(c^{\mathsmaller{(m)}}_{d,k} \mid \phi^{\mathsmaller{(m)}}_{d,k})P(\phi^{\mathsmaller{(m)}}_{d,k})\\
        &\propto_{\phi^{\mathsmaller{(m)}}_{d,k}} P(\textrm{y}^{\mathsmaller{(m)}}_{d,k} \mid  c^{\mathsmaller{(m)}}_{d,k}, \phi^{\mathsmaller{(m)}}_{d,k})  P(c^{\mathsmaller{(m)}}_{d,k})P(\phi^{\mathsmaller{(m)}}_{d,k})\\
&\propto_{\phi^{\mathsmaller{(m)}}_{d,k}} P(\textrm{y}^{\mathsmaller{(m)}}_{d,k} \mid  c^{\mathsmaller{(m)}}_{d,k}, \phi^{\mathsmaller{(m)}}_{d,k}) P(\phi^{\mathsmaller{(m)}}_{d,k})\\
&\propto_{\phi^{\mathsmaller{(m)}}_{d,k}} \textrm{Pois}(\textrm{y}^{\mathsmaller{(m)}}_{d,k};   c^{\mathsmaller{(m)}}_{d,k} \cdot \phi^{\mathsmaller{(m)}}_{d,k}) \Gamma(\phi^{\mathsmaller{(m)}}_{d,k}; c_0, d_0)\\
&\propto_{\phi^{\mathsmaller{(m)}}_{d,k}} \Gamma(\phi^{\mathsmaller{(m)}}_{d,k}; c_0 + \textrm{y}^{\mathsmaller{(m)}}_{d,k}, d_0  + c^{\mathsmaller{(m)}}_{d,k}) &\qed
\end{align}

\textbf{Complete conditional for $\boldsymbol{\left(\textrm{y}_{\obs,q}\right)_{q=1}^Q}$.} To derive the conditionals for the latent counts $\left(\textrm{y}_{\obs,q}\right)_{q=1}^Q$, we first observe that their sum $\yentry = \sum_{q=1}^Q \textrm{y}_{\obs, q}$ is observed. Due to the relationship between the Poisson and multinomial distributions, we have that 
\begin{align}
   \left(\left(\textrm{y}_{\obs,q}\right)_{q=1}^Q \mid - \right) &\sim  \textrm{Multi}(\yentry, \Pi_{\obs})\,,\\
   \Pi_{\obs} &\equiv \left(\frac{\lambda_q \prod_{m=1}^M \phi^{\mathsmaller{(m)}}_{d_m\kqm}}{\sum_{q'=1}^Q \lambda_{q'} \prod_{m=1}^M \phi^{\mathsmaller{(m)}}_{d_m\textrm{k}_{q',m}}}\right)_{q=1}^Q &\qed
\end{align}
where $\Pi_{\obs} \in \R^{Q}, \sum_{q=1}^Q \Pi_{\obs, q} = 1 \,, \Pi_{\obs, q} \geq 0$.

\textbf{Complete conditional for $\boldsymbol{\pi}^{\mathsmaller{\mathsmaller{(m)}}}$.}
By Dirichlet-multinomial conjugacy, for each $\boldsymbol{\pi}^{\mathsmaller{\mathsmaller{(m)}}}$, 
\begin{align}
    \left(\boldsymbol{\pi}^{\mathsmaller{\mathsmaller{(m)}}} \mid -\right) \, &\sim \hspace{0.5em} \textrm{Dirichlet}(\boldsymbol{\alpha}^{\mathsmaller{(m)}})\,, \hspace{1em} \boldsymbol{\alpha}^{\mathsmaller{(m)}} \in \R^{K_m},\\
    \textrm{where } \boldsymbol{\alpha}^{\mathsmaller{(m)}}_{k} &= \hspace{0.5em} \alpha_0 + \sum_{q=1}^Q 1\{\kqm = k\} &\qed
\end{align}
\textbf{Complete conditional for $\boldsymbol{\lambda_q}$.} Define $\textrm{y}_q$ as $\textrm{y}_q \equiv \sum_{\obs} \textrm{y}_{\obs, q}$, a sum over the latent counts $(\textrm{y}_{\obs, q})$.
\begin{align}
    P(\lambda_q \mid -) &= P(\lambda_q \mid \left(\Phi^{\mathsmaller{(m)}}\right)_{m=1}^M,\textrm{y}_{q}, (\textrm{k}_{q,m})_{m=1}^M)\\
    &\propto_{\lambda_q} P((\Phi^{\mathsmaller{(m)}})_{m=1}^M,\textrm{y}_{q}, (\textrm{k}_{q,m})_{m=1}^M \mid \lambda_q) P(\lambda_q)\\
    &= P(\lambda_q)P(\textrm{y}_{q} \mid (\Phi^{\mathsmaller{(m)}},(\textrm{k}_{q,m})_{m=1}^M, \lambda_q)P((\Phi^{\mathsmaller{(m)}})_{m=1}^M \mid (\textrm{k}_{q,m})_{m=1}^M, \lambda_q)P((\textrm{k}_{q,m})_{m=1}^M \mid \lambda_q)\\
    &= P(\lambda_q)P(\textrm{y}_{q} \mid (\Phi^{\mathsmaller{(m)}},\textrm{k}_{q,m})_{m=1}^M, \lambda_q)P((\Phi^{\mathsmaller{(m)}})_{m=1}^M)P((\textrm{k}_{q,m})_{m=1}^M)\\
    &\propto_{\lambda_q} P(\lambda_q)P(\textrm{y}_{q} \mid (\Phi^{\mathsmaller{(m)}},\textrm{k}_{q,m})_{m=1}^M, \lambda_q)\\
    &\propto_{\lambda_q}\Gamma(\lambda_q; a_0, b_0)\textrm{Pois}(\textrm{y}_{q}; \lambda_q \prod_{m=1}^M \sum_{d_m=1}^{D_m}\phi^{\mathsmaller{(m)}}_{d_m\kqm})\\
    &\propto_{\lambda_q} \lambda_q^{a_0 - 1}e^{-b_0 \lambda_q}e^{- \lambda_q \prod_{m=1}^M \sum_{d_m=1}^{D_m}\phi^{\mathsmaller{(m)}}_{d_m\kqm}}\lambda_q^{\textrm{y}_q}\\
    &\propto_{\lambda_q} \lambda_q^{a_0 + \textrm{y}_q - 1}e^{-\lambda_q[b_0 + \prod_{m=1}^M \sum_{d_m=1}^{D_m}\phi^{\mathsmaller{(m)}}_{d_m\kqm}]}\\
\text{implying that }\\
(\lambda_q \mid -) \, &\sim \Gamma\left(a_0 + \textrm{y}_q, b_0 + \prod_{m=1}^M \sum_{d_m=1}^{D_m}\phi^{\mathsmaller{(m)}}_{d_m\kqm}\right) &\qed
\end{align}

\textbf{Complete conditional for $\boldsymbol{\kqm}$.} We define
\begin{align}
\label{eq:app_kcond_start}
\textrm{y}_{{d_m \bullet}, q} = \sum_{d' \in \mathcal{D}, d'_m = d}\textrm{y}_{\obs', q} \sim \textrm{Pois}(\phi^{\mathsmaller{(m)}}_{d_m\textrm{k}_{q, m}} \underbrace{\lambda_q \sum_{d' \in \mathcal{D}, d'_m = d} \prod_{m' \neq m} \phi^{(m')}_{d'_{m'}\textrm{k}_{q, m'}}}_{c_{d_m, q}} )
\end{align}
Then 
\begin{align}
    P(\textrm{k}_{q,m} = k \mid -) &\propto_{\textrm{k}_{q,m}} \hspace{0.5em} P(\textrm{k}_{q,m} = k) P((\textrm{y}_{{d_m \bullet}, q})_{d_m = 1}^{D_m} \mid \textrm{k}_{q,m} = k, \lambda_q, \left(\textrm{k}_{q,m'}\right)_{m' \neq m}, \left(\Phi^{\mathsmaller{(m)}}\right)_{m=1}^M)\\
    &\propto_{\textrm{k}_{q,m}} \hspace{0.5em}\boldsymbol{\pi}^{\mathsmaller{\mathsmaller{(m)}}}_{k} \prod_{d_m=1}^{D_m}\textrm{Pois}(\textrm{y}_{{d_m \bullet}, q}; \phi^{\mathsmaller{(m)}}_{d_m, k} \lambda_q \sum_{d' \in \mathcal{D}, d'_m = d} \prod_{m' \neq m} \phi^{(m')}_{d'_{m'}\textrm{k}_{q, m'}})\\
    &\propto_{\textrm{k}_{q,m}} \hspace{0.5em}\boldsymbol{\pi}^{\mathsmaller{\mathsmaller{(m)}}}_{k} \prod_{d_m=1}^{D_m}\textrm{Pois}(\textrm{y}_{{d_m \bullet}, q}; \phi^{\mathsmaller{(m)}}_{d_m, k} c_{d_m, q})\\
    \label{eq:app_kcond_end}
    &= \hspace{0.5em} \frac{\boldsymbol{\pi}^{\mathsmaller{\mathsmaller{(m)}}}_{k} \prod_{d_m=1}^{D_m}\textrm{Pois}(\textrm{y}_{{d_m \bullet}, q}; \phi^{\mathsmaller{(m)}}_{d_m, k} c_{d_m, q})}{\sum_{k'=1}^{K_m} \boldsymbol{\pi}^{\mathsmaller{\mathsmaller{(m)}}}_{k'} \prod_{d_m=1}^{D_m}\textrm{Pois}(\textrm{y}_{{d_m \bullet}, q}; \phi^{\mathsmaller{(m)}}_{d_m, k'} c_{d_m, q})} &\qed
\end{align}

\subsection{Code for \allocore}
\href{https://github.com/jhood3/allocore}{https://github.com/jhood3/allocore}
\section{ADDITIONAL RESULTS}
\subsection{Synthetic Experiment}
 \label{sec:synthetic}
\begin{figure}[!htbp]
     \centering
     \includegraphics[width=0.9\linewidth]{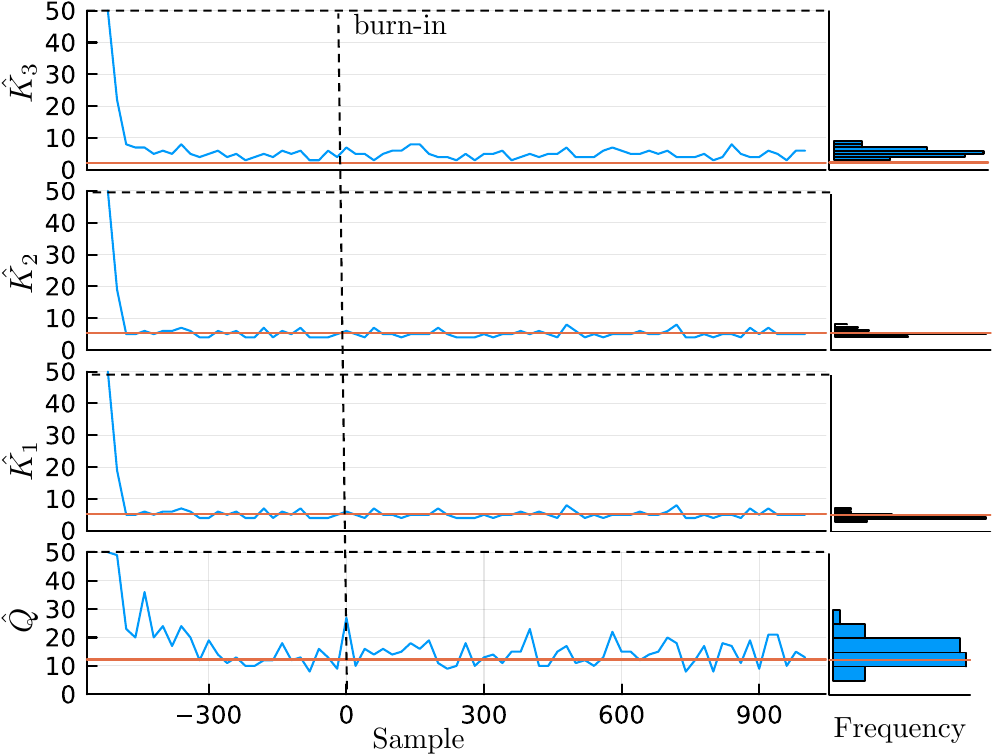}
     \caption{Posterior estimates of effective dimensionalities $K_1^*, K_2^*$ and $ K_3^*$, and number of classes $Q^*$ in a synthetic setting. \textit{Left:} Trace plot from the Gibbs sampler. \textit{Right:} Histograms of posterior samples. The red line denotes the ground truth value.}
    \label{fig:histTrace}
 \end{figure}

 The generative model defined for Bayesian Poisson \allocore~ implicitly defines a prior on the latent dimensionality $K_m$ of each mode $M$. As an exploratory exercise, we generate synthetic data to evaluate if ~\allocore~can detect ground-truth latent dimensions $\left(K_m^*\right)_{m=1}^M$ and $||\Lambda^*||_0$ in the simplest of settings. Our synthetic design generates $\Ytensor$ according to Bayesian Poisson~\allocore, 
 \begin{align}
\yentry = \sum_{q=1}^Q\textrm{y}_{\obs,q},\hspace{1em} \textrm{y}_{\obs,q} \stackrel{\textrm{ind.}}\sim \Pois{\widehat{\textrm{y}}_{\obs,q}},\hspace{1em}
\widehat{\textrm{y}}_{\obs,q} &\equiv \lambda_q \prod_{m=1}^M \phimdk{m}{\im}{\kqm}
\end{align}
 Factor matrices $\Phi^{(1)}, \Phi^{(2)}$ are sampled column-wise as
   $\boldsymbol{\phi}^{(1)}_{:, \textrm{k}_1},\, \boldsymbol{\phi}^{(2)}_{:, \textrm{k}_2} \sim \textrm{Dirichlet}(\mathbf{0.01})$
and scaled such that each column sums to $5$ (each entry is multiplied by 5). We require heterogeneity over the columns of $\Phi^{(3)}$, fixing $\boldsymbol{\phi}^{(3)}_{:, 1} = (2.5, 0.5, 1, 0.5, 0.5)$, $\boldsymbol{\phi}^{(3)}_{:, 2} = (0.5, 2.5, 1, 0.5, 0.5)$.
Setting $K_m = Q = 50$, we fit~\allocore~to $\Ytensor$. Figure~\ref{fig:histTrace} shows the trajectory of sampled $\{(K_m^{(s)})_{m=1}^3, Q^{(s)}\}^S_{s=1}$ during training. \allocore~ infers core tensors with effective shape approximate to that of $\boldsymbol{\Lambda}^*$. Histograms over samples estimate the marginal posterior distribution for each $K_m$ and $Q$. While~\allocore~overestimates $K_3^*$, the posterior samples concentrate around $K_1^*, K_2^*$, and $Q^*$, indicating recovery of ground truth parameters. We leave for future work theoretical investigation these properties of~\allocore. 
\vfill
 \pagebreak
 
\subsection{Quantitative Results}
\begin{figure}[h!]
    \begin{center}
\resizebox{\textwidth}{!}{%
\includegraphics[height=3cm]{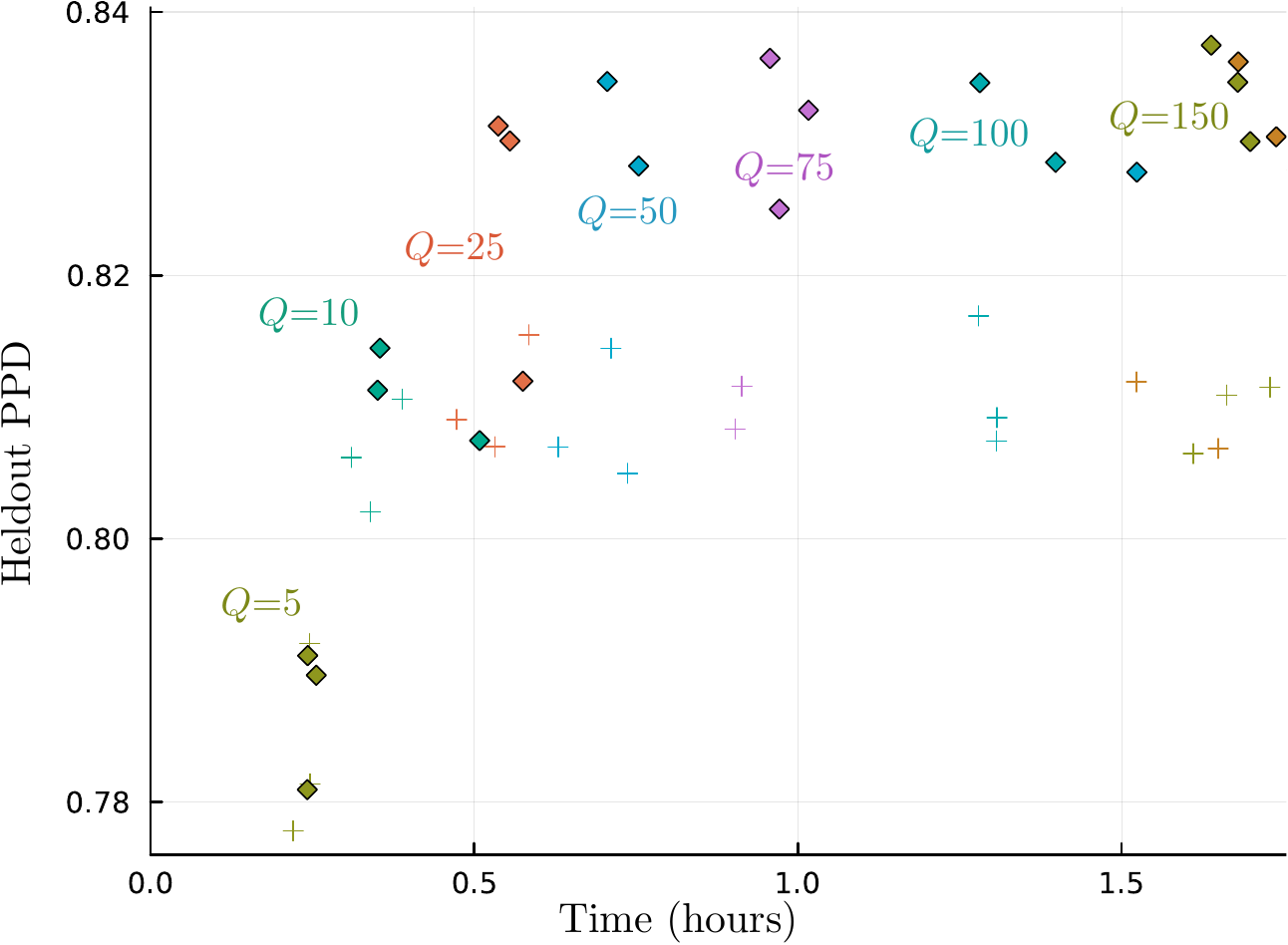}%
\quad
\includegraphics[height=3cm]{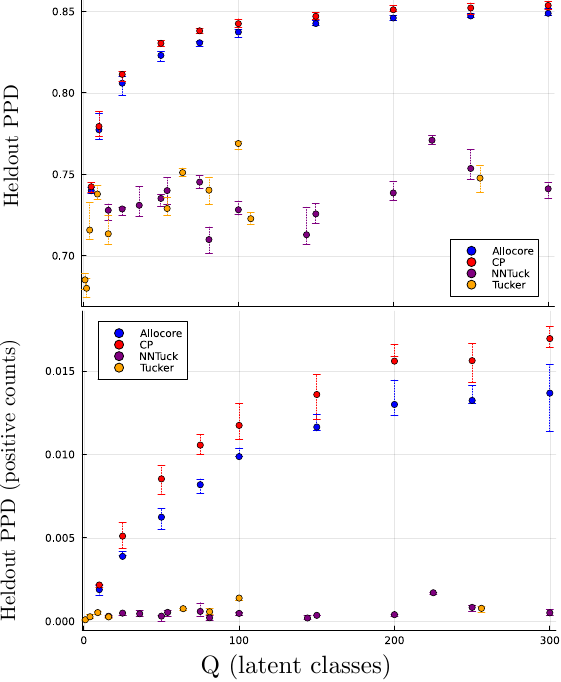}%
}
\end{center}
           \caption{\textit{Left:} PPD by wall-clock time (ICEWS data). Each point is an~\allocore~model run for 5,000 iterations, with a unique combination of $Q$, denoted by color, and core tensor size, denoted by $\diamond$- (small) versus $+$-shaped (large) markers. Performance plateaus early suggesting that sparse $\allocore$~achieves the same performance as full Tucker at only a fraction of the cost. \textit{Right:} Heldout PPD on the full heldout set (top) and the heldout set restricted to positive counts (bottom) across $Q$. \allocore~is plotted in blue, CP in red, NNTuck in purple, and Tucker in yellow. Error bars span the interquartile range across masks.}
           \label{icewsq}
\end{figure}

Supplementary quantitative results on the ICEWS dataset are given in~\cref{icewsq}.

\subsection{Qualitative Results}
\label{sec:app_qualitative}
We plot each learned latent class for an \allocore~model fitted to ICEWS monthly data, a tensor of size $249 \times 249 \times 20 \times 228$, with $Q = 400,$ $C = 50, D = 50$, $K = 20$, and $R = 300$, ordered by the non-zero core values $\lambda_{\lat}$ in descending order. 
\vfill
\pagebreak

\begin{figure}[!htbp]
\centering
\foreach \n in {1,...,8} {
    \includegraphics[height=0.22\textheight]{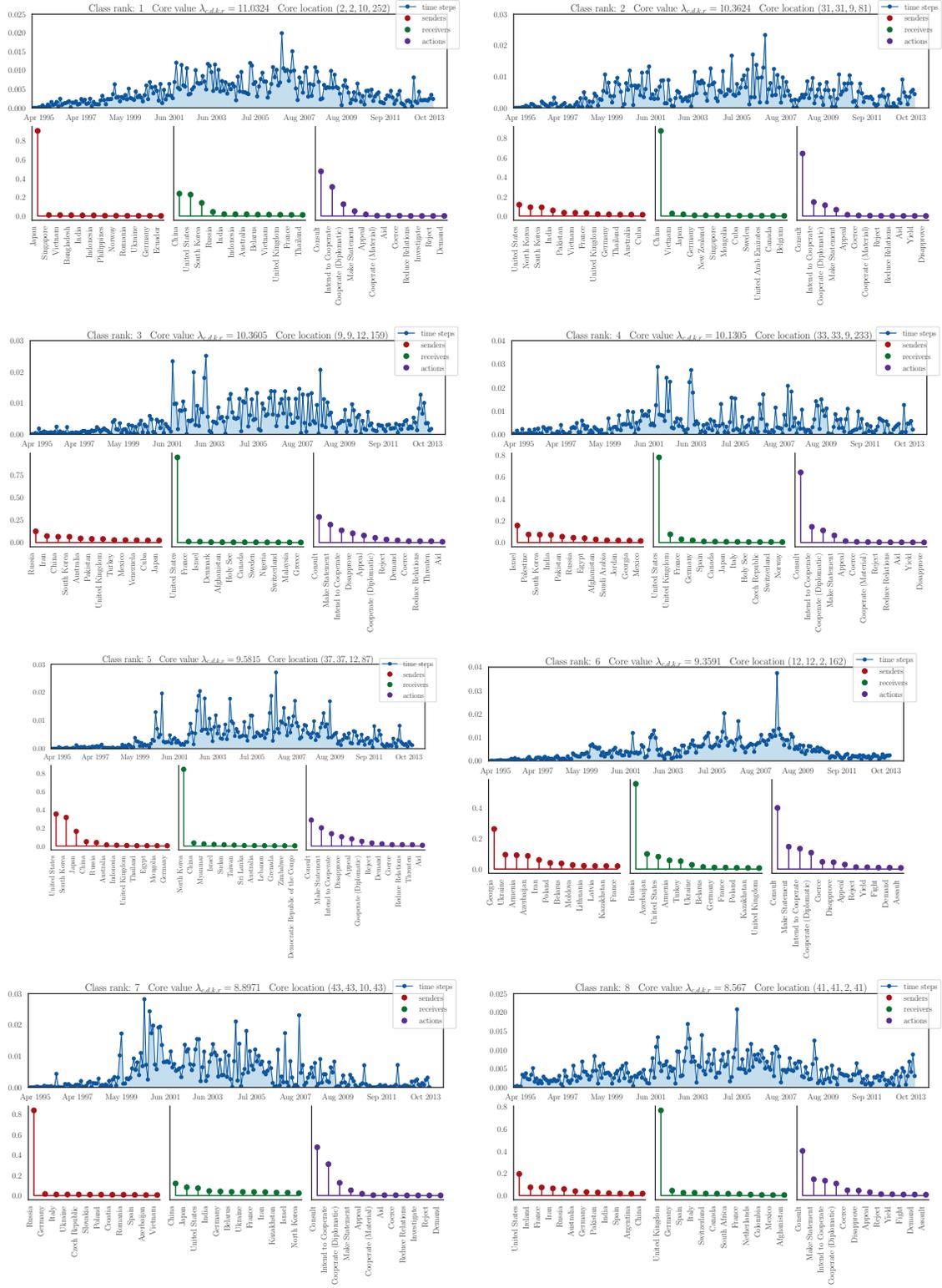}
    \ifnum\n<8\relax\else\caption{Inferred latent classes 1--8.}\label{fig:class1}\fi
}
\end{figure}

\foreach \x in {9,17,...,100} { 
    \begin{figure}[!htbp]
    \centering
    \foreach \n in {\x,...,\the\numexpr\x+7\relax} {
        \includegraphics[height=0.22\textheight]{figures/component_figs/component_\n.pdf}
        \ifnum\n<\the\numexpr\x+7\relax\relax\else\caption{Inferred latent classes \x--\the\numexpr\x+7\relax.}\label{fig:class\x}\fi 
    }
    \end{figure}
}

\end{document}